\documentclass[lettersize,journal]{IEEEtran}
\usepackage{amsmath,amsfonts}
\usepackage{algorithmic}
\usepackage{algorithm}
\usepackage{array}
\usepackage[caption=false,font=normalsize,labelfont=sf,textfont=sf]{subfig}
\usepackage{textcomp}
\usepackage{stfloats}
\usepackage{url}
\usepackage{verbatim}
\usepackage{graphicx}
\usepackage{cite}
\usepackage{svg}
\usepackage{amssymb}
\usepackage[percent]{overpic} 
\usepackage{booktabs}
\usepackage{multirow}
\usepackage{hyperref}
\usepackage{colortbl}   
\usepackage{xcolor}
\usepackage{makecell}
\usepackage{enumitem}
\usepackage{pifont}
\newcommand{\ie}{\textit{i}.\textit{e}.}
\newcommand{\eg}{\textit{e}.\textit{g}.}
\newcommand{\conf}[1]{\textcolor{gray}{\scriptsize (\textit{#1})}}
\newcommand{\model}{WARM-CAT}
\definecolor{tabcolor}{HTML}{FFF0F5}
\begin{document}


\title{\model \includegraphics[width=0.9em]{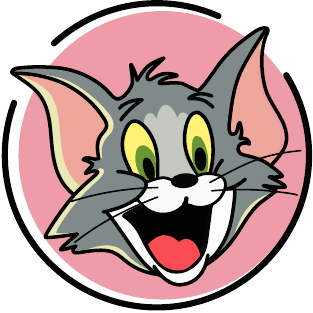}: Warm-Started Test-Time Comprehensive Knowledge Accumulation for Compositional Zero-Shot Learning}


\author{Xudong Yan, Songhe Feng, Jiaxin Wang, Xin Su, Yi Jin
\thanks{Xudong Yan, Songhe Feng, Jiaxin Wang, Xin Su, and Yi Jin are with the School of Computer Science and Technology, Beijing Jiaotong University, Beijing 100044, China (e-mail: xud\_yan@bjtu.edu.cn; shfeng@bjtu.edu.cn; jiax-wang@bjtu.edu.cn; xinsue@bjtu.edu.cn; yjin@bjtu.edu.cn)}
\thanks{Manuscript received December 19, 2025;}
\thanks{Corresponding author: Songhe Feng.}}

\markboth{}%
{\model: Warm-Started Test-Time Comprehensive Knowledge Accumulation for Compositional Zero-Shot Learning}

\IEEEpubid{0000--0000/00\$00.00~\copyright~2021 IEEE}

\maketitle

\begin{abstract}
Compositional Zero-Shot Learning (CZSL) aims to recognize novel attribute-object compositions based on the knowledge learned from seen ones. 
Existing methods suffer from performance degradation caused by the distribution shift of label space at test time, which stems from the inclusion of unseen compositions recombined from attributes and objects. 
To overcome the challenge, we propose a novel approach that accumulates comprehensive knowledge in both textual and visual modalities from unsupervised data to update multimodal prototypes at test time. Building on this, we further design an adaptive update weight to control the degree of prototype adjustment, enabling the model to flexibly adapt to distribution shift during testing. Moreover, a dynamic priority queue is introduced that stores high-confidence images to acquire visual prototypes from historical images for inference. 
Since the model tends to favor compositions already stored in the queue during testing, we warm-start the queue by initializing it with training images for visual prototypes of seen compositions and generating unseen visual prototypes using the mapping learned between seen and unseen textual prototypes.
Considering the semantic consistency of multimodal knowledge, we align textual and visual prototypes by multimodal collaborative representation learning. 
To provide a more reliable evaluation for CZSL, we introduce a new benchmark dataset, C-Fashion, and refine the widely used but noisy MIT-States dataset. 
Extensive experiments indicate that our approach achieves state-of-the-art performance on four benchmark datasets under both closed-world and open-world settings. 
The source code and datasets are available at \url{https://github.com/xud-yan/WARM-CAT}.
\end{abstract}

\begin{IEEEkeywords}
Compositional zero-shot learning, compositionality, test-time adaptation, label space shift.
\end{IEEEkeywords}

\section{Introduction}
\label{section1:introduction}
\begin{figure}[t]
    \centering
    \includegraphics[width=0.98\linewidth]{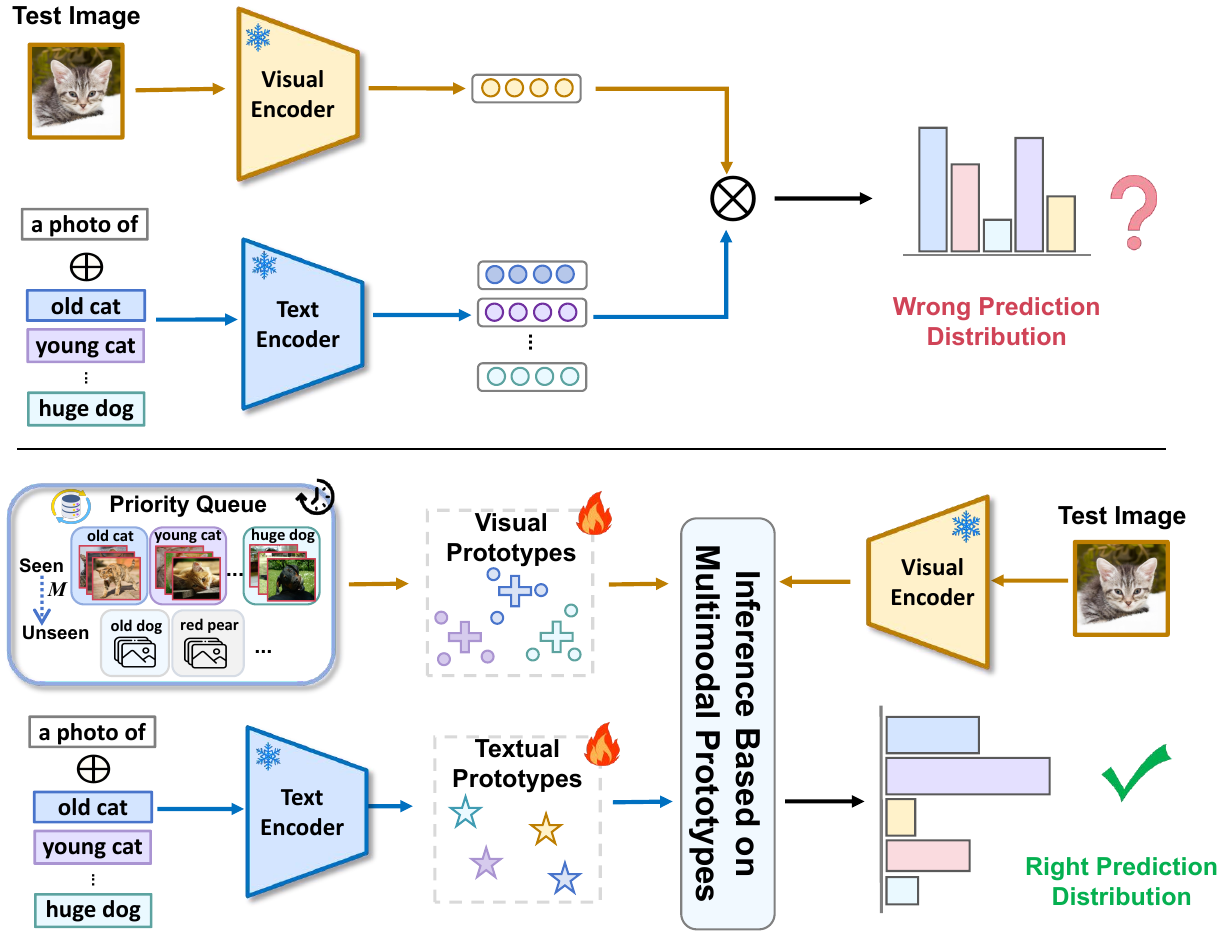}
    \caption{\textbf{At test time}, existing methods (\textit{top}) fail to adapt using test images, resulting in biased prediction distributions due to label space shift. By contrast, \model\ (\textit{bottom}) progressively accumulates multimodal knowledge from unsupervised test data, enabling effective adaptation to address this challenge.}
    \label{fig:vice_figure}
\end{figure}
\IEEEPARstart{C}{ombining} existing concepts to understand and recognize novel compositions embodies a fundamental aspect of human intelligence ~\cite{1948_Philo_human1,2013_spring_human2,2017_BBS_human3}. 
Even for compositions that have never been encountered, such as \texttt{wilted sunflower} or \texttt{ivory wolf}, humans can effortlessly infer their appearance by recombining known attributes and objects drawn from prior experience.
This capability reflects the broader principle of compositionality and generalization, through which a limited set of primitives (\ie, attributes and objects) supports an expansive range of conceptual compositions~\cite{2016_arxiv_general,2020_human_4,2018_arxiv_human5}.
Motivated by this core aspect of human intelligence, Compositional Zero-Shot Learning (CZSL) has emerged to learn representations of attributes and objects from seen attribute-object pairs and leverage the knowledge to recognize unseen compositions~\cite{2017_CVPR_first, 2019_ICCV_TMN,2020_NIPS_noise_causal, 2020_CVPR_SymNet, 2021_CVPR_Compcos_open}. 
Beyond its conceptual appeal, CZSL addresses a practical challenge in visual recognition: categories defined as holistic units scale poorly under combinatorial growth, while attributes and objects recur across contexts and can be reused in a compositional manner~\cite{2017_CVPR_first}. By modeling this structure explicitly, CZSL alleviates the dependence on exhaustive annotations and improves robustness under increasingly novel attribute–object configurations. 

\IEEEpubidadjcol

Traditional CZSL approaches can be broadly divided into two main streams based on whether attribute–object pairs are disentangled in the visual space. 
One main stream aims at learning representations of compositions by aligning images and textual labels in a shared space, and predicting compositions with the lowest distance~\cite{2021_CVPR_CGE_CGQA, 2022_PAMI_CoCGE, 2023_WACV_atten_propa}. 
The other stream concentrates on disentangling visual representations of compositions and predicting individual primitives separately to reduce composition learning into primitive learning since both seen and unseen compositions inherently share the same attributes and objects~\cite{2021_NIPS_ProtoProp, 2022_CVPR_SCEN, 2023_CVPR_ADE}. 
Recently, benefiting from the impressive multimodal representational abilities of large pre-trained vision-language models (VLMs) like CLIP~\cite{2021_ICML_CLIP}, there have been further significant advancements in CZSL
~\cite{2023_ICLR_CSP, 2023_CVPR_DFSP, 2024_CVPR_Troika}. 
These methods employ the prompt tuning technique~\cite{2022_CVPR_Coop}, \ie, replacing the hard prompt like "\textit{a photo of} \texttt{[attribute] [object]}" with learnable soft prompt tokens to align images and textual composition labels, thereby fine-tuning VLMs to better suit the CZSL task. 

While these pioneering methods have achieved significant progress, they fail to address the performance degradation due to the distribution shift of label space 
at test time. 
Specifically, models are assigned to identify unseen attribute-object compositions that are absent during training. This pattern leads to a mismatch between the learned and actual test-time label distribution, bringing about inaccurate predictions and poor generalization. 
The key reason why the above issue has yet to be effectively addressed lies in the fact that the model parameters and class prototypes are frozen after training, preventing the model from leveraging the test data to adapt towards the new distribution. 

However, in more real-world scenarios, an outstanding intelligent system should continuously accumulate knowledge and thus overcome distribution changes after deployment by utilizing the unlabeled data provided through user interactions during the testing phase. 
In CZSL, the following three aspects should be taken into consideration when using test-time unsupervised data to adjust the model: 
(1) \textbf{Accumulation. }The process of observing test-time images can be regarded as a form of knowledge accumulation, rather than a complete adaptation to unseen compositions at the cost of forgetting knowledge of seen compositions learned during training~\cite{2017_PNAS_forget1, 2022_ICML_forget2_time2}. 
(2) \textbf{Knowledge Comprehensiveness. }Existing methods treat compositions encoded by the text encoder of VLMs as textual-modal prototypes, and predict compositions by computing similarity between the visual features of images and prototypes. However, they overlook the potential of utilizing visual information from the historical images encountered at test time. 
(3) \textbf{Efficiency. }Considering the practical scenario of interaction with users, the method should be time-efficient with low latency \cite{2022_ICML_forget2_time2,2023_ICCV_time1}.

To this end, we propose a \textbf{W}arm-st\textbf{AR}ted test-ti\textbf{M}e \textbf{C}omprehensive knowledge \textbf{A}ccumula\textbf{T}ion (\textbf{\model}) approach for CZSL, a novel framework that leverages multimodal knowledge of unlabeled data at test time to overcome label distribution shift, as illustrated in Fig.~\ref{fig:vice_figure}. 
During the test phase, we keep the textual-modal prototypes frozen and progressively learn a Knowledge Accumulation Module (KAM) for prototype adjustment to bridge the label distribution gap with the continual influx of test samples. 
Subsequently, we determine the extent to which KAM updates the prototypes by employing a designed adaptive update weight strategy,  
based on similarities between the image and the prototypes. 
To take full advantage of the visual knowledge from previously seen images, \model\ maintains a dynamic priority queue to store high-confidence images for each class. 
Since the model tends to predict compositions that already exist in the priority queue during testing, we warm-start the queue by initializing it with training images of seen compositions. Then, we propose to apply the mapping relationship learned between seen and unseen textual prototypes to the visual prototypes of seen compositions to generate virtual visual prototypes for unseen compositions. 
Building on this, visual-modal prototypes are constructed from the images stored in the queue and are dynamically updated\textendash similarly to the textual counterpart\textendash by visual KAM with higher-confidence images enqueued. 
As testing progresses, the textual and visual prototypes are updated by the entropy minimization objective~\cite{2021_ICLR_Tent,2022_NIPS_TPT}, and are jointly used to facilitate composition recognition under the new label distribution. 
In addition, given the inherent semantic interdependence between multimodal knowledge, we align textual and visual prototypes by multimodal collaborative representation learning. 

For evaluating CZSL, the fashion domain epitomizes the need for compositional reasoning, as novel styles arise from compositions of clothing categories and fine-grained visual attributes such as color, material, and pattern.
However, suitable benchmarks for this domain remain conspicuously absent, despite its central role in online shopping and recommendation systems.
To fill this gap, we construct C-Fashion, a compositional reasoning dataset built upon FashionIQ~\cite{2021_CVPR_FashionIQ}. 
In addition, since the widely used MIT-States~\cite{2015_CVPR_mit} suffers from substantial noise, with about 70\% of its labels being incorrect~\cite{2020_NIPS_noise_causal,2022_ECCV_INV,2025_IJCAI_Trident}, we clean and refine this dataset for fairer evaluation.
We further propose evaluation metrics tailored to long-tailed distributions in CZSL and conduct a systematic analysis of existing methods under this setting.
Extensive experiments on four benchmark datasets (\ie, UT-Zappos~\cite{2014_CVPR_ut_zappos}, our newly proposed C-Fashion, C-GQA~\cite{2021_CVPR_CGE_CGQA}, and the refined MIT-States$^*$) demonstrate that our \model\ outperforms the state-of-the-art by significant margins in both closed-world and open-world settings (Sec.~\ref{section3_1:task_formulation}).

In summary, the contributions of our work are five-fold:
\begin{itemize}[leftmargin=*,noitemsep]
    \item We propose \model, a novel framework that accumulates multimodal knowledge and updates prototypes from unlabeled data at test time to bridge the label distribution shift. To the best of our knowledge, we are the first to leverage unsupervised data at test time to improve model performance in CZSL.
    \item \model\ adopts a priority queue that stores historical high-confidence images to calculate visual prototypes and adaptively updates multimodal prototypes by knowledge accumulation modules.
    \item We propose a new benchmark dataset C-Fashion for compositional reasoning in fashion domain, and refine the noisy dataset MIT-States. 
    \item We introduce novel evaluation metrics to assess CZSL under long-tailed distributions and present a systematic analysis of existing methods.
    \item Extensive experiments conducted on four benchmark datasets demonstrate that our \model\ achieves state-of-the-art performance in both closed-world and open-world settings.
\end{itemize}

This paper significantly extends our preliminary work~\cite{2019_NIPS_TOMCAT} published in NeurIPS 2025 in many aspects. In this extended version, several substantial improvements are concluded here: 1) We warm-start the priority queue by initializing it with images of seen compositions, and by generating virtual unseen visual prototypes through the mapping relationship learned between seen and unseen textual prototypes, thereby preventing the model from being biased towards the compositions of historical images. 2) We propose a new benchmark dataset C-Fashion, filling the gap of CZSL task lacking fashion domain benchmark. 3) A refined version of MIT-States is provided by relabeling the images, resulting in a more reliable benchmark for CZSL evaluation. 4) We introduce new evaluation metrics designed for assessing CZSL under long-tailed distributions and provide a comprehensive analysis of current baseline methods. These extensions enhances the model’s stability, ensures fairer comparisons, and enables more comprehensive experiments and in-depth analysis.
\section{Related Work}
\label{section2:related_work}

\textbf{Compositional Zero-Shot Learning (CZSL).} 
CZSL requires the model to recognize unseen compositions with the attribute and object knowledge learned from seen compositions. 
Prior works can be broadly categorized according to whether attribute–object pairs are disentangled in the visual space.
On the one hand, composition-based approaches model the intrinsic interactions between attributes and objects without separating their visual contributions~\cite{2017_CVPR_first, 2018_ECCV_AttrAsOp, 2019_ICCV_TMN, 2021_CVPR_CGE_CGQA, 2022_PAMI_CoCGE, 2023_WACV_atten_propa}. 
These methods typically embed images and textual compositions into a shared space. 
AttOp~\cite{2018_ECCV_AttrAsOp} encodes attribute-induced transformations of objects and imposes constraints such as inverse consistency, commutativity, and antonym relations. 
SymNet~\cite{2020_CVPR_SymNet} enforces symmetry by jointly using a disentangling and a coupling module, allowing it to apply and remove attributes in a consistent manner. 
Graph-based approaches further build relational structures among primitives and compositions, leveraging graph neural networks to propagate information from seen to unseen nodes~\cite{2021_CVPR_CGE_CGQA,2022_MM_CVGAE,2022_PAMI_CoCGE}. 
On the other hand, vision-disentangling approaches explicitly learn primitive-level visual representations to reduce composition learning into primitive learning since both seen and unseen compositions inherently share the same attributes and objects
~\cite{2021_NIPS_ProtoProp,2022_PAMI_Symnet2,2022_CVPR_OADis,2023_CVPR_CANET, 2024_PAMI_SAD_SP, 2024_IJCAI_CCZSL}. 
These methods align attribute and object embeddings in separate visual spaces before recombining them. 
Representative directions include conditional attribute modeling based on object identity~\cite{2023_CVPR_CANET}, attention-based control of primitive interactions~\cite{2022_TMM_Attention1,2023_CVPR_ADE}, triplet-based learning using informative image triplets~\cite{2022_CVPR_SCEN,2022_CVPR_OADis,2025_IJCAI_Trident}, and domain-invariance constraints applied to embeddings or gradients~\cite{2022_ECCV_INV,2025_TPAMI_DBC}. 
Rather than learning image-composition association from scratch, recent approaches have increasingly shifted focus to exploiting the multimodal representational capacity of VLMs (\eg, CLIP~\cite{2021_ICML_CLIP}) for CZSL~\cite{2023_ICLR_CSP,2023_CVPR_DFSP,2024_CVPR_Troika,2024_ECCV_PLID,2024_CVPR_CDS}. 
Troika~\cite{2024_CVPR_Troika} exploits multi-path paradigm and cross-modal traction modules to jointly model attributes, objects, and compositions. 
CDS-CZSL~\cite{2024_CVPR_CDS} leverages context-based diversity-driven specificity learning to prioritize specific attributes with richer information. 
ClusPro~\cite{2025_ICLR_ClusPro} learns multiple clustering‑based prototypes for each attribute primitive, dynamically discovering and updating them via within‑primitive clustering. 
PLO~\cite{2025_MM_PLO} employs a pre-trained LLM to generate progressive visual cues and uses them to guide CLIP through a step-by-step visual analysis, enabling a progressively hierarchical understanding of compositions.
Although significant progress has been achieved, these methods freeze model parameters and prototypes after training, preventing them from leveraging test-time data for further improvement.

\textbf{Vision-Language Model (VLM).}
VLMs, such as CLIP~\cite{2021_ICML_CLIP} and ALIGN~\cite{2021_ICML_ALIGN}, are pre-trained on large-scale web image-text datasets and have demonstrated remarkable capability in learning joint representations that align visual and textual modalities. This cross-modal alignment enables VLMs to perform a wide range of downstream tasks, including image classification, retrieval, object detection, and compositional reasoning. For example, CLIP~\cite{2021_ICML_CLIP} leverages a contrastive learning objective to associate images with their corresponding text descriptions, achieving general zero-shot performance without task-specific supervision. 
To repurpose the pre-trained multimodal knowledge of VLMs for downstream tasks, fine-tuning paradigms are proposed that are parameter-efficient and task-adaptive. On the textual side, \emph{prompt tuning} has emerged as an effective strategy, where manually crafted prompts are replaced with learnable soft tokens while keeping the text encoder frozen~\cite{2022_CVPR_Coop,2022_CVPR_cocoop}. This technique significantly reduces the number of trainable parameters and allows the model to adapt to new tasks without overfitting. CoOp~\cite{2022_CVPR_Coop} introduces prompt learning under few-shot settings, eliminating the reliance on hand-crafted prompts. CoCoOp~\cite{2022_CVPR_cocoop} further extends this idea by conditioning the prompts on instance-specific features, which enhances generalization to unseen classes. 
On the visual side, \emph{adapter tuning} has become a popular approach, wherein lightweight trainable modules, such as residual adapters, bottleneck MLPs, or attention-based adapters, are inserted into the visual backbone while keeping the majority of parameters frozen~\cite{2023_CVPR_TaskRes,2023_EMNLP_apollo,2024_IJCV_CLIP_adapter,2025_TMM_MVP_qu}. This allows efficient adaptation to downstream tasks without full fine-tuning of the backbone. 
AdapterFormer~\cite{2022_NIPS_AdapterFormer} augments each Transformer block in the visual encoder with a parallel adapter, fusing the outputs via element-wise addition, which improves task-specific performance while maintaining the pre-trained representation quality. 
TaskRes~\cite{2023_CVPR_TaskRes} integrates residual adapters into ResNet-based visual encoders for task-specific tuning. MVP-CLIP~\cite{2025_TMM_MVP_qu} proposes multi-view visual adapters to capture complementary visual information for complex tasks.
In this work, following CoOp~\cite{2022_CVPR_Coop} and AdapterFormer~\cite{2022_NIPS_AdapterFormer}, we fine-tune CLIP using training data to obtain a simple yet effective base model for subsequent testing.

\textbf{Online Test-time Adaptation (OTTA). }
OTTA refers to a practical technique where a trained model
continually adapts itself by exploiting a stream of unsupervised test data\textendash each test sample is used exactly once, and the model is required to retain and leverage the knowledge gained from earlier test images to improve its performance on later ones in an online manner
~\cite{2021_ICLR_Tent,2022_CVPR_CoTTA, 2023_ICLR_sar_tta,2024_ICLR_vida_tta,2025_IJCV_TTA_Survey}. 
Tent~\cite{2021_ICLR_Tent} proposes to optimize batch normalization layer by minimizing prediction entropy on test data. 
CoTTA~\cite{2022_CVPR_CoTTA} is the first work to decrease the accumulation error and avoid catastrophic forgetting through using averaged pseudo-labels and retaining the knowledge of the original model to enhance long-term adaptation. 
SAR~\cite{2023_ICLR_sar_tta} enhances generalization ability in OTTA by eliminating partially noisy samples with high gradients and promoting the convergence of model weights to a flat minimum. 
ViDA~\cite{2024_ICLR_vida_tta} introduces a homeostatic knowledge allotment strategy to fuse low-rank and high-rank features, and thus improves distinct domain representations. 
Recently, activating the zero-shot capability of VLMs at test time to mitigate domain shift in downstream tasks has increasingly attracted significant research attention
~\cite{2022_NIPS_TPT,2024_CVPR_TDA,2024_NIPS_DPE,2024_NIPS_boost, 2024_NIPS_historical, 2025_WACV_TPS}. 
For instance, 
TDA~\cite{2024_CVPR_TDA} proposes a training-free dynamic adapter that uses a lightweight key-value cache and progressive pseudo-label refinement without backpropagation.
DynaPrompt~\cite{2025_ICLR_DynaPrompt} adaptively selects and optimizes the relevant prompts for each test data built on an online prompt buffer. 
TPS~\cite{2025_WACV_TPS} proposes a straightforward and efficient prototype shifting approach that adjusts per-class prototypes within the embedding space.
Notably, OTTA mainly focuses on addressing distribution shifts in the feature domain while we aim to resolve the challenge of label distribution shift caused by unseen compositions recombined from attributes and objects in CZSL. 
Similar to OTTA, CZSL does not have access to any data of unseen compositions during training.

\begin{figure}[t]
    \centering
    \includegraphics[width=0.9\linewidth]{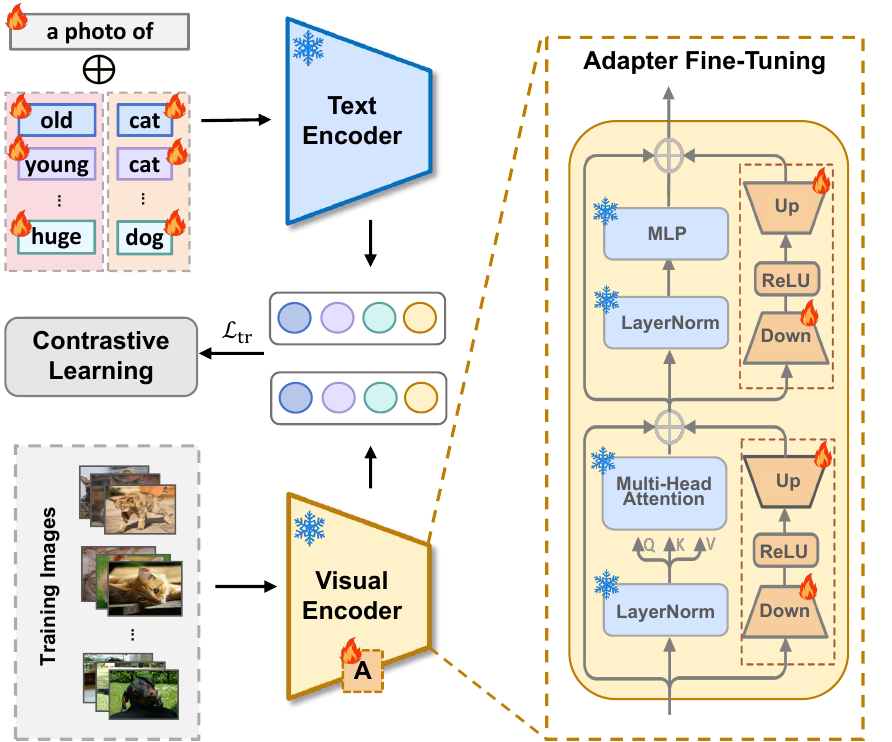}
    \caption{Prompt tuning of the text encoder and adapter tuning of the visual encoder during training.}
    \label{fig:adapter}
\end{figure}

\section{Method}
\label{section3:method}
\begin{figure*}[t]
    \centering
    \begin{overpic}
    [width=0.99\textwidth]
    {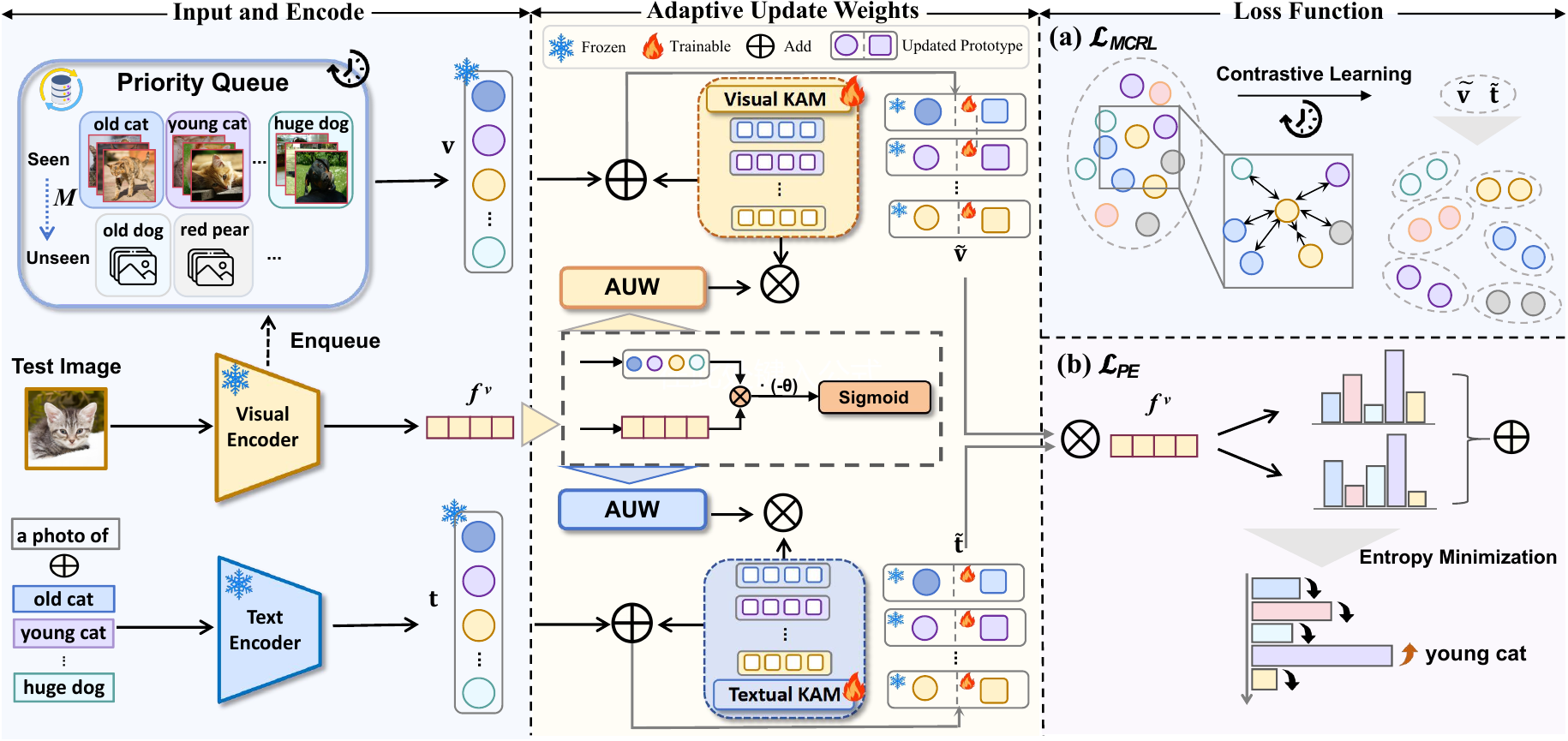}
    \end{overpic}
    \caption{The overall architecture of our proposed \model\ at test time. The model accumulates multimodal knowledge to update prototypes to overcome the label distribution shift.}
    \label{fig:main_figure}
\end{figure*}
\subsection{Task Formulation}
\label{section3_1:task_formulation}
In this section, we provide a formal description of the CZSL task. Given an attribute set $A$ and an object set $O$, the composition set can be defined by the Cartesian product of $A$ and $O$, \ie, $C = A \times O$. And the seen and unseen composition set $C^s$ and $C^u$ are two disjoint subsets of $C$, \ie, $C^s \cap C^u = \varnothing$. During the training phase, the model learns from a seen training set $D^{tr} = \{(x, c) | x \in \mathcal{X}, c \in C^s\}$, where $\mathcal{X}$ is the image space, and $c$ is the composition label of image $x$. In the closed-world setting~\cite{2019_ICCV_TMN}, the test composition set $C^{te}$ is defined as the union of $C^s$ and $C^u$, \ie, $C^{te} = C^s \cup C^u$, where only the presupposed known composition space is considered. For the open-world setting~\cite{2021_CVPR_Compcos_open}, the test composition set expands to all possible attribute-object pairs, \ie, $C^{te} = C$.

\subsection{Training Phase}
\label{section3_2:training_phase}
As a preparatory step, we train a simple CLIP-based model by adjusting only its visual and text encoders on the training set, which serves as the foundation for subsequent testing. 

\textbf{Textual Representation Extraction. } CLIP treats each attribute-object composition label as a prompt-based textual input, \ie, "\textit{a photo of} \texttt{[attribute] [object]}", which is subsequently encoded via the text encoder $\psi$. Given a composition $c = (a, o)$, we create a learnable prompt $\mathbf{P}^c = [\mathbf{p}_1, \dots, \mathbf{p}_l, \mathbf{w}^a, \mathbf{w}^o]$, where $\mathbf{p}_{1:l}$ represents the learnable prompt tokens. $\mathbf{w}^a$ and $\mathbf{w}^o$ are the learnable word tokens of $a$ and $o$ within the attribute and object vocabularies, respectively. Therefore, the textual representation of composition $c$ is obtained, \ie, $f^t_c = \psi(\mathbf{P}^c)$.

\textbf{Visual Representation Extraction. }Given an input image $x \in \mathbb{R}^{H \times W \times 3}$, we feed it into the CLIP visual encoder $\phi$ and employ the output \texttt{[CLS]} token as its visual representation, \ie, $f^v = \phi(x)$. Based on AdapterFormer~\cite{2022_NIPS_AdapterFormer}, a set of learnable adapters is injected into the visual encoder of CLIP, while keeping the initial parameters frozen. As shown in Fig.~\ref{fig:adapter}, given the input embedding $e$ of each layer, the adapter is formulated as:
\begin{equation}
    A(e) = \mathbf{W}^{\text{up}}(ReLU(\mathbf{W}^{\text{down}} e))\ ,
\end{equation}
where $\mathbf{W}^{\text{down}}$ and $\mathbf{W}^{\text{up}}$ are learnable linear layers employed for down-sampling and up-sampling, respectively. For each layer, the final output is obtained by adding the adapter output to the original output through a residual connection.

\textbf{Training Objective. }Given the textual and visual representations, we compute the probability $p(c|x)$ and use the cross-entropy loss to align them:
\begin{equation}
\begin{alignedat}{2}
&\mathcal{L}_{tr} = -\log p(c \mid x), \\[0.5ex]
&p(c \mid x) = \frac{\exp(\cos(f^v, f^t_c)/\tau)}
{\sum_{c^{\prime} \in C^{s}} \exp(\cos(f^v, f^t_{c^{\prime}})/\tau)},
\end{alignedat}
\end{equation}
where $\tau$ denotes the temperature parameter of pre-trained CLIP and $\cos(\cdot, \cdot)$ is used to compute cosine similarity. After the training phase, 
a simple yet effective base model is obtained by only tuning prompt and adapters within CLIP, without introducing any additional complex modules to accelerate inference.

\subsection{\model\ at Test Time}
\label{section3_3:\model}
As the major novelty, we introduce \model\ to overcome the challenge of label distribution shift by employing unsupervised data at test time, as shown in Fig.~\ref{fig:main_figure}. Specifically, textual and visual composition prototypes are obtained by CLIP-encoded textual labels and a dynamic priority queue of historical images, respectively. These multimodal prototypes are then updated by Knowledge Accumulation Modules (KAMs) and adaptive update weights, with the objective of minimizing prediction entropy at test time.

\textbf{Textual Prototype Construction. }In line with CLIP's principles, we treat the embeddings of both seen and unseen composition labels\textendash encoded by the text encoder of the base model after training\textendash as the textual-modal prototypes, \ie, $\mathbf{t} = [\mathbf{t}_{c_1}, \mathbf{t}_{c_2}, \dots, \mathbf{t}_{c_{|C^{te}|}}]^\top \in \mathbb{R}^{|C^{te}| \times d}$.

\textbf{Visual Prototype Construction. }Inspired by TDA~\cite{2024_CVPR_TDA} and DPE~\cite{2024_NIPS_DPE}, complementary to the text modality, we recognize that prior visual knowledge\textendash captured from historical test images\textendash can be leveraged to further enhance the discriminative capability of CLIP. Therefore, a dynamic priority queue is designed to selectively preserve $K$ ($K=3$) high-confidence images, which enables \model\ to retain representative and reliable exemplars throughout the test phase. Specifically, for each seen and unseen composition $c$, we maintain a "\textit{confidence-feature}" queue, \ie, $q^c = [(h_i, f^v_i)]^{K}_{i=1}$, where $f^v_i \in \mathbb{R}^d$ is the visual feature of the image $x_i$ in the queue and $h_i \in \mathbb{R}$ is its prediction entropy as confidence:
\begin{equation}
\begin{alignedat}{2}
    &\mathcal{H}(x_i) = - \sum_{c \in C^{te}} p(c|x_i, \Tilde{\mathbf{t}}) \log p(c|x_i, \Tilde{\mathbf{t}})\ , \\
    &p(c|x_i, \Tilde{\mathbf{t}}) = \frac{\exp( \cos(f^v_i, \Tilde{\mathbf{t}}_c) / \tau)}{\sum_{c^{\prime} \in C^{te}} \exp(\cos(f^v_i, \Tilde{\mathbf{t}}_{c^{\prime}}) / \tau)}\ ,
\end{alignedat}
\end{equation}
where $\Tilde{\mathbf{t}}$ is the updated textual prototypes (defined below in Eq.~\ref{equ:t_wave}). The lower prediction entropy means higher confidence, and the samples in the queue are sorted by their confidence, \ie, $h^c_i \leq h^c_{(> i)}$. Based on the priority queue, the visual-modal prototype for each composition is computed by averaging the visual features, \ie, $\mathbf{v}_c = \frac{1}{K} \sum_{i=1}^K f_i^v$. 
Subsequently, the visual prototypes are denoted as $\mathbf{v} = [\mathbf{v}_{c_1}, \mathbf{v}_{c_2}, \dots, \mathbf{v}_{c_{|C^{te}|}}]^\top \in \mathbb{R}^{|C^{te}| \times d}$.

\textbf{Priority Queue Warm-Start. }If the priority queue is initially empty, the model tends to predict compositions that have already been stored during testing. Therefore, we offline warm-start the queue by initializing it once after training using visual features. Specifically, for each seen composition $c^s \in C^s$, we extract the visual features of its training images $f^v$ and compute their prediction entropy $h$. The top-K elements $[(h_i, f^v_i)]^{K}_{i=1}$ with the lowest prediction entropy are stored for initialization of the priority queue corresponding to their label.

\begin{figure}[t]
    \centering
    \includegraphics[width=0.98\linewidth]{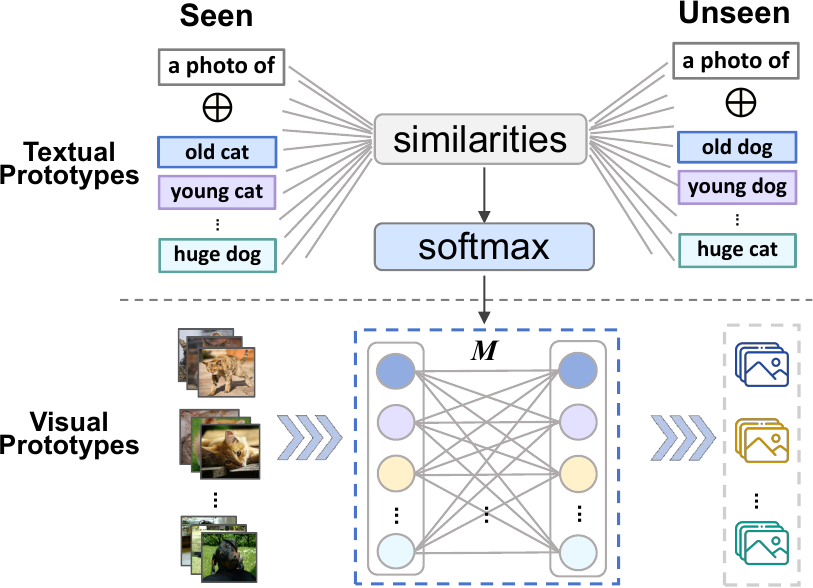}
    \caption{Learning the mapping relationship between seen and unseen textual prototypes (\textit{top}) and applying it to seen visual prototypes to obtain unseen visual prototypes (\textit{bottom}).}
    \label{fig:mapping}
\end{figure}

For unseen compositions, the training images are missing, so their visual prototypes cannot be directly obtained at first. However, we assume that the semantic relationships among compositions should ideally remain consistent across both visual and textual spaces~\cite{2013_NIPS_devise,2021_ICML_CLIP}. Therefore, virtual visual prototypes are generated for unseen compositions by applying the mapping learned between seen and unseen textual prototypes to the visual prototypes of seen compositions in Fig.~\ref{fig:mapping}. Specifically, for seen and unseen textual prototypes $\mathbf{t}^s \in \mathbb{R}^{|C^{s}| \times d}$ and $\mathbf{t}^u \in \mathbb{R}^{|C^{u}| \times d}$, the mapping relationship is obtained in a training-free manner:
\begin{equation}
    M = \text{Softmax}(\cos(\mathbf{t}^s, \mathbf{t}^u) / \tau_M) \in \mathbb{R}^{|C^{s}| \times |C^{u}|}\ ,
\end{equation}
where $\tau_M$ is the temperature parameter, and $M$ denotes the cosine-similarity-based mapping matrix that transfers semantic information from seen compositions to unseen ones. The mapping matrix can be applied to seen visual prototypes $\mathbf{v}^s \in \mathbb{R}^{|C^{s}| \times d}$ to obtain unseen visual prototypes:
\begin{equation}
    \mathbf{v}^u = M^T\mathbf{v}^s \in \mathbb{R}^{|C^{u}| \times d} \ .
\end{equation}
The generated visual prototypes for unseen compositions are inserted as visual features into the priority queue with an extreme value taken as the prediction entropy.

\textbf{Priority Queue Update. }For an incoming  image $x$, we first calculate its visual feature $f^v$ and the prediction entropy $h$, and obtain its pseudo-label $c_p$ by assigning the composition with the highest predicted probability, \ie, $c_p = \arg \max_{c \in C^{te}} p(c|x, \Tilde{\mathbf{t}})$. 
If the queue for $c_p$ is not full, the pair ($h$, $f^v$) is directly inserted into the queue. If the queue is full, only if the new image has a prediction entropy lower than the highest element in the queue, the highest one is replaced with ($h$, $f^v$); otherwise, the queue remains unchanged. 

\textbf{Knowledge Accumulation Module (KAM). }We aim to continuously acquire information about the new distribution from test samples, while avoiding both catastrophic forgetting of existing knowledge and excessive increases in inference latency. To this end, we introduce learnable KAMs instead of directly modifying the parameters of the base model, and employ adaptive update weights to control the extent to which the KAMs adjust the prototypes. Specifically, for multimodal prototypes $\mathbf{t}$ and $\mathbf{v}$, multimodal KAMs consist of two sets of learnable parameters, which are initialized to zero:
\begin{equation}
\begin{aligned}
    \Delta\mathbf{t} = [\Delta\mathbf{t}_{c_1}, \Delta\mathbf{t}_{c_2}, \dots, \Delta\mathbf{t}_{c_{|C^{te}|}}]^\top \in \mathbb{R}^{|C^{te}| \times d}\ ,\\
    \Delta\mathbf{v} = [\Delta\mathbf{v}_{c_1}, \Delta\mathbf{v}_{c_2}, \dots, \Delta\mathbf{v}_{c_{|C^{te}|}}]^\top \in \mathbb{R}^{|C^{te}| \times d}\ .
\end{aligned}
\end{equation}

\textbf{Adaptive Update Weight.} Taking textual prototypes as an example, we demonstrate the process by which the prototypes are updated with a newly arrived image $x$. Specifically, the cosine similarity is computed between the visual feature $f^v$ of the image and each original prototype $\textbf{t}_c$, based on which the adaptive update weight is calculated as follows:
\begin{equation}
    w_c = \sigma(-\theta\cdot s_c)\ ,\ \ \ s_c = \cos(f^v, \mathbf{t}_c)\ ,
\end{equation}
where $\sigma$ denotes the Sigmoid activation function and $\theta$ is the hyperparameter that controls the degree of update. Therefore, the updated textual prototypes can be denoted as follows:
\begin{equation}
   \Tilde{\mathbf{t}} = [\Tilde{\mathbf{t}}_{c_1}, \Tilde{\mathbf{t}}_{c_2}, \dots, \Tilde{\mathbf{t}}_{c_{|C^{te}|}}]^{\top}\ , \ \ \ \Tilde{\mathbf{t}}_c = \frac{\mathbf{t}_c + w_c\Delta\mathbf{t}_c}{||\mathbf{t}_c + w_c\Delta\mathbf{t}_c||}\ .
   ~\label{equ:t_wave}
\end{equation}

Accordingly, we can obtain the updated visual prototypes $\Tilde{\mathbf{v}}$. This adaptive weighting mechanism enables more controlled update of the prototypes, thereby avoiding treating all compositions equally regardless of familiarity. Intuitively, when the test image closely matches the original prototype, it is likely associated with a seen composition, and thus excessive adjustments should be avoided. Conversely, a large discrepancy between the test image and the prototype suggests a potentially unseen composition, permitting stronger update to improve adaptability.

\textbf{Prediction Entropy Minimization. }Following Tip-Adapter~\cite{2022_ECCV_tip_adapter}, the final prediction for the input image is determined as follows:
\begin{equation}
\begin{aligned}
    p(c|x, \Tilde{\mathbf{t}}, \Tilde{\mathbf{v}}) = \frac{\exp{(f^v \cdot \Tilde{\mathbf{t}}_c + \alpha \mathcal{A}(f^v, \Tilde{\mathbf{v}}_c))}}{\sum_{c^\prime \in C^{te}} \exp{(f^v \cdot \Tilde{\mathbf{t}}_{c^\prime} + \alpha \mathcal{A}(f^v, \Tilde{\mathbf{v}}_{c^\prime}))}}\ ,\\[1ex]
    \mathcal{A}(f^v, \Tilde{\mathbf{v}}_c) = \exp{(- \beta (1-f^v \cdot \Tilde{\mathbf{v}}_c))}\ ,
    \label{eq:prediction_probability}
\end{aligned}
\end{equation}
where $\alpha$ and $\beta$ are hyperparameters controlling multimodal balance and modulating visual-modal sharpness, respectively. At test time, minimizing prediction entropy serves as an unsupervised learning signal that encourages the model to produce more confident predictions in the target label space, thereby enhancing generalization. By reducing uncertainty, the model progressively adjusts to better aligning with the test distribution, which includes unseen compositions. The loss for multimodal prediction entropy is formulated as follows:
\begin{equation}
    \mathcal{L}_{PE} = - \textstyle \sum_{c \in C^{te}} p(c|x, \Tilde{\mathbf{t}}_c, \Tilde{\mathbf{v}}_c) \log{p(c|x, \Tilde{\mathbf{t}}_c, \Tilde{\mathbf{v}}_c)}.
    \label{eq:lpe}
\end{equation}

\textbf{Multimodal Collaborative Representation Learning. }Considering the intrinsic semantic interdependence between multimodal knowledge, we align the textual and visual prototypes by employing multimodal collaborative representation learning. This strategy effectively facilitates the integration of both modalities, 
thereby enhancing the representation of both textual and visual information in a unified framework. Specifically, contrastive learning is exploited to bring the visual and textual prototypes corresponding to the same composition closer while pushing apart non-corresponding ones, as formulated below: 
\begin{equation}
\begin{aligned}
\mathcal{L}_{MCRL}
= -\frac{1}{2|C^{te}|}\sum_{c\in C^{te}}\Bigg(
    &\log{\frac{\exp{(\cos(\Tilde{\mathbf{t}}_c, \Tilde{\mathbf{v}}_c) /  \tau)}}%
    {\sum_{c^\prime\in C^{te}}\exp{(\cos(\Tilde{\mathbf{t}}_c, \Tilde{\mathbf{v}}_{c^\prime})/\tau)}}} \\
    & \!\!\!\!\!\!\!\!\!\!\!\!+ \log{\frac{\exp{(\cos(\Tilde{\mathbf{t}}_c , \Tilde{\mathbf{v}}_c)/\tau)}}%
    {\sum_{c^\prime\in C^{te}}\exp{(\cos(\Tilde{\mathbf{t}}_{c^\prime}, \Tilde{\mathbf{v}}_{c})/\tau)}}}
\Bigg).
\label{eq:lmcrl}
\end{aligned}
\end{equation}

\textbf{Testing Pipeline Overview. }Upon receiving a test image, we first extract its visual feature and compute the prediction entropy with the original textual prototypes. We then determine whether to update the priority queue of its pseudo-composition label based on the entropy. After performing multi-modal prototype refinement by KAMs and adaptive update weights, the final inference prediction for this image is obtained. To reduce latency, the backpropagation update of KAMs is deferred until after the inference step by minimizing the total loss of multimodal prediction entropy and multimodal collaborative representation learning as follows:
\begin{equation}
    \mathcal{L}_{\model} = \mathcal{L}_{PE} + \lambda\mathcal{L}_{MCRL}\\ ,
\end{equation}
where $\lambda$ is the weighting coefficient of $\mathcal{L}_{MCRL}$.

\begin{table*}[t]
    \centering
    \caption{Summary statistics of the four datasets used in our experiments. }
    \resizebox{0.9\linewidth}{!}{
        \begin{tabular}{llccclcclccclccc}
        \toprule
        \multirow{2}{*}{Dataset}&&\multicolumn{3}{c}{\textit{Composition}}&&\multicolumn{2}{c}{\textit{Train}}&&\multicolumn{3}{c}{\textit{Validation}}&&\multicolumn{3}{c}{\textit{Test}} \\\cmidrule{3-5}\cmidrule{7-8} \cmidrule{10-12} \cmidrule{14-16}
          && $|A|$ & $|O|$ & $|A \times O|$ && $|C^s|$ & $|\mathcal{X}|$ && $|C^s|$ & $|C^u|$ & $|\mathcal{X}|$ && $|C^s|$ & $|C^u|$ & $|\mathcal{X}|$ \\ \midrule
        UT-Zappos~\cite{2014_CVPR_ut_zappos} && 16 & 12 & 192 && 83 & 22998 && 15 & 15 & 3214 && 18 & 18 & 2914 \\
        C-Fashion && 76 & 28 & 2128 && 350 & 19018 && 60 & 59 & 5221 && 60 & 58 & 5466 \\
        C-GQA~\cite{2021_CVPR_CGE_CGQA} && 413 & 674 & 278362 && 5592 & 26920 && 1252 & 1040 & 7280 && 888 & 923 & 5098 \\ 
        MIT-States$^*$ &&  141 & 266 & 37506 && 1074 & 9851 && 200 & 186 & 2085 && 201 & 184 & 2143 \\
        
        \bottomrule
        \end{tabular}
    }
    
    \label{tab:dataset_split}
\end{table*}

\section{Experiment}
\label{section4:Experiment}

\subsection{C-Fashion Benchmark Dataset Construction}
\begin{figure}[t]
    \centering
    \includegraphics[width=0.98\linewidth]
    {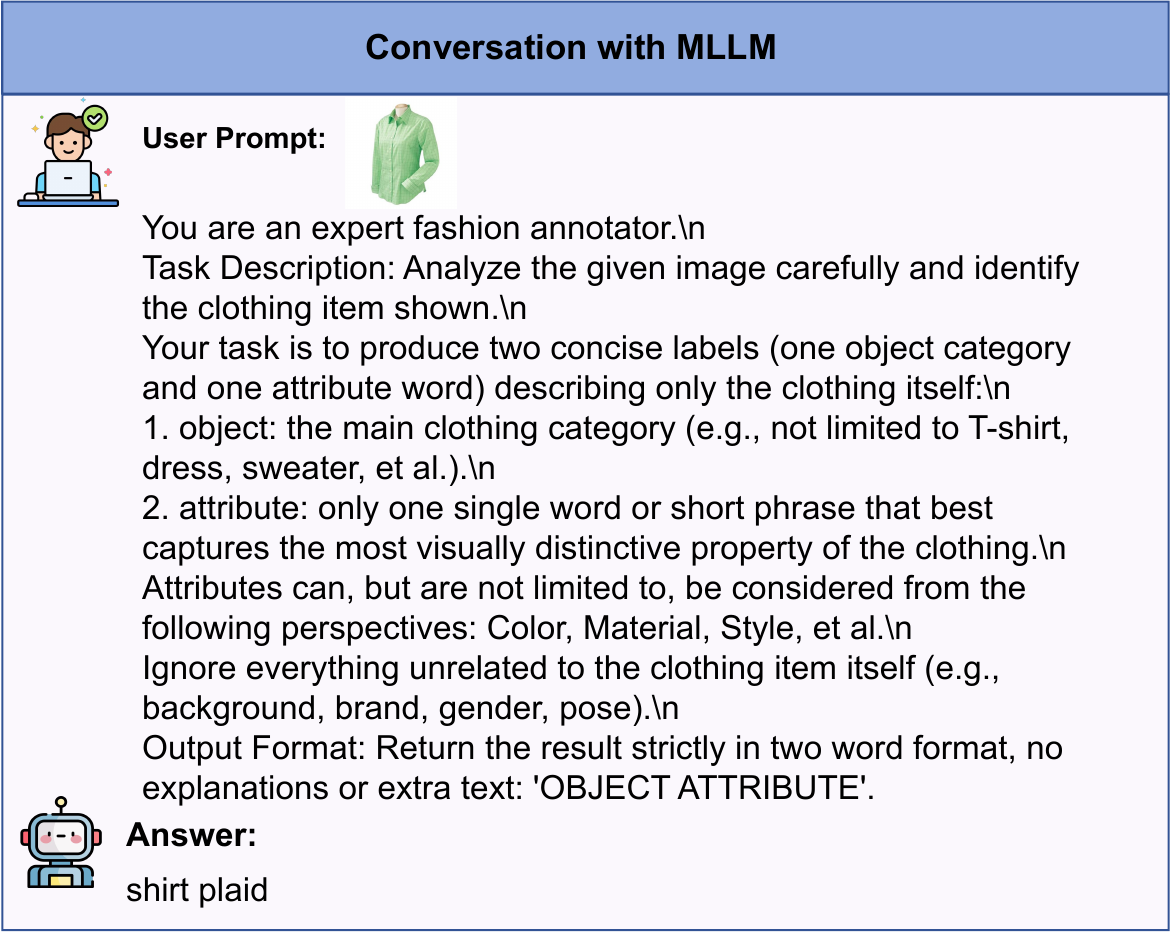} 
    \caption{Image annotation generated by MLLM for C-Fashion benchmark dataset.}
    \label{fig:annotate}
\end{figure}
The fashion domain inherently relies on compositional reasoning over clothing categories and fine-grained visual attributes, but tailored benchmarks for this setting are still missing despite its practical significance in online shopping and recommendation systems.
Therefore, we introduce a new benchmark dataset C-Fashion to fill the gap that CZSL lacks fashion domain benchmark. C-Fashion is proposed based on the original FashionIQ~\cite{2021_CVPR_FashionIQ}, a widely used benchmark for interactive and attribute-based image retrieval. Based on the original FashionIQ images, we performed a new round of annotation and filtration for compositional reasoning. 

\textbf{Annotation and Filtering. }As shown in Fig.~\ref{fig:annotate}, the images and prompt texts are first provided to InternVL-3-8b~\cite{2025_arxiv_internvl3} to generate initial annotations. Since MLLM-based labeling is not error-free, the subsequent filtering step is required.
The correction and filtering procedures are carried out sequentially as follows: 
1) images containing garbled text or irrelevant characters are removed; 
2) case, characters and related formats are standardized according to the CZSL dataset conventions; 
3) objects, attributes and compositions with only few associated images are discarded, as extremely sparse categories hinder reliable training and lead to unstable evaluation; 
4) images exhibiting low similarity to their assigned labels are filtered out; 
5) for compositions with an excessive number of images, a subset is randomly discarded.

\textbf{Data Split.}
To construct the training, validation, and test partitions, 
we first reorganize the dataset into seen and unseen composition splits. Given all images annotated with compositional labels, seen and unseen compositions are sampled at a ratio of 3:1, while ensuring that the seen compositions contain about five times as many images as the unseen ones. To guarantee that the model observes the full primitives during training, the set of seen compositions is further expanded so that the training split covers all objects and all attributes. 
For seen compositions, we allocate three-quarters of them to the training set, and divide the remaining ones equally into validation and test subsets. 
For unseen compositions, whose images are not allowed to appear in training, we split their images evenly between the validation and test sets, thus enabling a fair assessment of generalization to novel compositions. 
This protocol ensures both comprehensive semantic coverage during training and a clean separation between seen and unseen compositions during evaluation.

As a result, the training split comprises 350 composition pairs with 19k images covering 76 objects and 28 attributes. The validation split includes 60 seen and 59 unseen pairs totaling 5k images, while the test split contains 60 seen and 58 unseen pairs with 5k images.

\subsection{MIT-States Benchmark Dataset Refinement}
MIT-States~\cite{2015_CVPR_mit} contains diverse real-world objects and attributes collected by early image search engine technology. However, given the substantial noise in MIT-States, with approximately 70\% of the labels being incorrect~\cite{2020_NIPS_noise_causal, 2022_ECCV_INV, 2025_IJCAI_Trident}, we conduct a systematic refinement of the dataset following the same protocol used for constructing C-Fashion and propose MIT-States$^*$. 
Consequently, the training split contains 1,074 composition pairs and 10k images spanning 266 objects and 141 attributes. The validation split comprises 201 seen and 184 unseen pairs with a total of 2k images, whereas the test split includes 200 seen and 186 unseen pairs, also amounting to 2k images. Compared with the original dataset, the refined version contains a larger set of compositions, implying a broader search space. Despite providing fewer training images, models achieve better performance on the new version, indicating that the refined dataset is less noisy and of higher quality.

\subsection{Experiment Setup}
\label{section4_1:Experiment_setup}

\textbf{Datasets.}
Our proposed \model\ is evaluated on four challenging datasets: UT-Zappos~\cite{2014_CVPR_ut_zappos}, our newly proposed C-Fashion, C-GQA~\cite{2021_CVPR_CGE_CGQA}, and the refined MIT-States$^*$. 
UT-Zappos is a fine-grained fashion dataset consisting of different shoes (\eg, \texttt{heels}, \texttt{sandals}) with their material (\eg, \texttt{leather}, \texttt{satin}). 
For the low-noise dataset C-Fashion, the objects are mainly various kinds of clothing (\eg, \texttt{dress}, \texttt{blouse}), while the attributes are their properties(\eg, \texttt{black}, \texttt{sleeveless}). 
C-GQA~\cite{2021_CVPR_CGE_CGQA} is a large-scale dataset composed of in-the-wild images with more general compositions (\eg, \texttt{marble countertop}, \texttt{striped pillow}). 
MIT-States$^*$ is a comprehensive dataset that captures a wide spectrum of real-world objects and their associated attributes. 
Table~\ref{tab:dataset_split} reports the common data splits of the four datasets. 
UT-Zappos~\cite{2014_CVPR_ut_zappos} contains 16 attributes and 12 objects for 29,126 images, and it is split into 83 seen and 15/18 unseen compositions for training and validation/testing, respectively. 
C-Fashion contains 29,705 images, with
76 attributes and 28 objects. It includes 350 seen compositions and 59/58 (validation/test) unseen compositions.
C-GQA~\cite{2021_CVPR_CGE_CGQA} is a natural image dataset which contains 39,298 images, with 413 attributes and 764 objects. It includes 5,592 seen compositions and 1,040/923 (validation/test) unseen compositions.
MIT-States$^*$ contains 14079 images, with 141 attributes and 266 objects. It comprises 1,074 seen compositions
and 184/186 (validation/test) unseen compositions.

\textbf{Metrics.}
Following the evaluation protocol of previous works~\cite{2019_ICCV_TMN, 2021_CVPR_Compcos_open, 2023_ICLR_CSP}, a bias term from $-\infty$ to $+\infty$ is introduced to trade off the prediction logits between seen and unseen compositions. 
A positive bias value indicates improved prediction accuracy on unseen compositions, whereas a negative value suggests that the model is inclined toward seen compositions. 
By varying the bias term, we can draw a curve by taking the classification accuracy of seen compositions as the $x$-axis and that of the unseen compositions as the $y$-axis. 
To assess the overall performance on both seen and unseen compositions, we calculate the Area Under the Curve (\textbf{AUC}) and identify the point that yields the highest Harmonic Mean (\textbf{HM}) of the two accuracies. 
We also report the best \textbf{Seen} and \textbf{Unseen} accuracies obtained by setting the bias to $-\infty$ and $+\infty$, respectively. 
In the open-world setting, a post-training feasibility calibration is applied to filter out infeasible compositions within a vast search space~\cite{2021_CVPR_Compcos_open}.

\begin{table*}[t]
    \centering
    \caption{Closed-world results on UT-Zappos, C-Fashion, C-GQA, and MIT-States$^*$. The best results are displayed in \textbf{boldface}, and the second-best results are \underline{underlined}. The four indicators are explained in Metrics (Sec.~\ref{section4_1:Experiment_setup}).}
    \resizebox{\linewidth}{!}{
    \begin{tabular}
    {l>{\columncolor{tabcolor}}cccc>{\columncolor{tabcolor}}cccc>{\columncolor{tabcolor}}cccc>{\columncolor{tabcolor}}cccc}
    
        \toprule
         \multirow{2}{*}{Methods} & \multicolumn{4}{c}{UT-Zappos} & \multicolumn{4}{c}{C-Fashion} & \multicolumn{4}{c}{C-GQA} & \multicolumn{4}{c}{MIT-States$^*$}\\
         
         \cmidrule(lr){2-5} \cmidrule(lr){6-9} \cmidrule(lr){10-13} \cmidrule(lr){14-17}
         
          & AUC & HM & Seen & Unseen & AUC & HM & Seen & Unseen & AUC & HM & Seen & Unseen & AUC & HM & Seen & Unseen\\
         \midrule

         
         CLIP~\cite{2021_ICML_CLIP} \conf{ICML'21} & 5.0 & 15.6 & 15.8 & 49.1 & 11.8 & 25.7 & 27.0 & 59.4 & 1.4 & 8.6 & 7.5 & 25.0 & 19.6 & 36.6 & 35.3 & 64.2 \\  
         
         CoOp~\cite{2022_CVPR_Coop} \conf{IJCV'22} & 18.8 & 34.6 & 52.1 & 49.3 & 39.1 & 53.7 & 67.7 & 64.6 & 4.4 & 17.1 & 20.5 & 26.8 & 24.7 & 40.5 & 39.2 & 72.0 \\ 
         
         CSP~\cite{2023_ICLR_CSP} \conf{ICLR'23} & 33.0 & 46.6 & 64.2 & 66.2 & 39.4 & 54.9 & 66.0 & 65.7 & 6.2 & 20.5 & 28.8 & 26.8 & 27.6 & 44.2 & 44.3 & 70.3 \\ 
         
         DFSP~\cite{2023_CVPR_DFSP} \conf{CVPR'23} & 36.0 & 47.2 & 66.7 & 71.7 & 44.0 & 58.2 & 69.0 & 70.2 & 10.5 & 27.1 & 38.2 & 32.0 & 31.4 & 46.7 & 46.2 & 75.8 \\ 
    
         Troika~\cite{2024_CVPR_Troika} \conf{CVPR'24} & 41.7 & 54.6 & 66.8 & 73.8 & 47.5 & 59.3 & 74.3 & 72.8 & 12.4 & 29.4 & 41.0 & 35.7 & 36.3 & 50.6 & 49.7 & 80.2 \\ 
         
         PLID~\cite{2024_ECCV_PLID} \conf{ECCV'24} & 38.7 & 52.4 & 67.3 & 68.8 & 43.0 & 54.8 & \underline{76.3} & 64.3 & 11.0 & 27.9 & 38.8 & 33.0 & 31.5 & 47.8 & 45.6 & 76.1 \\ 

         RAPR~\cite{2024_AAAI_RAPR} \conf{AAAI'24} & 44.5 & 56.5 & 69.4 & 72.8 & 39.7 & 53.1 & 64.8 & 66.8 & 14.4 & 32.0 & 45.6 & 36.0 & 34.2 & 49.2 & 47.9 & 78.8 \\ 

         IMAX~\cite{2025_TPAMI_IMAX} \conf{TPAMI'25} & 40.6 & 54.2 & 69.3 & 70.7 & 40.5 & 55.7 & 64.5 & 67.7 & 12.8 & 29.8 & 39.7 & 35.8 & 36.7 & 52.0 & 49.3 & \textbf{82.1} \\ 

         ClusPro~\cite{2025_ICLR_ClusPro} \conf{ICLR'25} & 46.6 & 58.5 & 70.7 & \textbf{76.0} & 47.9 & 59.5 & 73.5 & \textbf{74.6} & 14.9 & 32.8 & 44.3 & 37.8 & 36.8 & 52.0 & 50.2 & 80.1 \\ 

        TOMCAT~\cite{2019_NIPS_TOMCAT} \conf{NeurIPS'25} & \underline{48.3} & \underline{60.2} & \underline{74.5} & 72.8 & \underline{48.0} & \underline{59.7} & 74.3 & \underline{72.9} & \underline{16.0} & \underline{34.0} & \underline{45.3} & \underline{40.1} & \underline{39.2} & \underline{52.7} & \textbf{55.0} & 79.7 \\ 
        
         \midrule
         
         \textbf{\model\ (Ours)} & \textbf{52.9} & \textbf{64.3} & \textbf{78.4} & \underline{73.9} & \textbf{51.5} & \textbf{63.6} & \textbf{76.6} & 72.5 & \textbf{16.9} & \textbf{35.4} & \textbf{45.6} & \textbf{41.5}  & \textbf{39.4} & \textbf{52.8} & \underline{54.1} & \underline{80.8} \\ 
         
        \bottomrule
    \end{tabular}
    }
    
    \label{tab:cw_results}
\end{table*}

\textbf{Implementation Details.}
We implement the base model with CLIP ViT-L/14 architecture in the training phase and \model\ at test time in PyTorch~\cite{2019_NIPS_pytorch} framework. The trainable prompt of CLIP ViT-L/14 in the training phase is initialized by "\textit{a photo of}". The learning rate of WARM-CAT is set to 5e-6, 5e-6, 1e-5, 1e-5 for UT-Zappos, C-Fashion, C-GQA, and MIT-States$^*$, respectively. The number of images in priority queue $K$ is 3 across all four datasets. 
For the hyperparameters that control cross-modal weighting and refine the sharpness of the visual modality ($\alpha, \beta$), we set them to (1, 15), (0.0625, 2.5), (0.5, 5), (1, 3.75), respectively. 
$\theta$ is set to 1, 3, 3, 1.5, and 
$\lambda$ is set to 3.5, 1.25, 2.75, 2.5 for the respective datasets.
The source code and datasets are also released at \href{https://github.com/xud-yan/WARM-CAT}{this website} to provide all implementation details and thus facilitate reproducibility. 

\textbf{Baselines.} We compare our \model\ with recent and prominent CLIP-based approaches on the four benchmarks, including CLIP~\cite{2021_ICML_CLIP}, CoOp~\cite{2022_CVPR_Coop}, CSP~\cite{2023_ICLR_CSP}, DFSP~\cite{2023_CVPR_DFSP}, 
Troika~\cite{2024_CVPR_Troika}, PLID~\cite{2024_ECCV_PLID}, RAPR~\cite{2024_AAAI_RAPR}, IMAX~\cite{2025_TPAMI_IMAX}, ClusPro~\cite{2025_ICLR_ClusPro}, and TOMCAT~\cite{2019_NIPS_TOMCAT}. 
CLIP~\cite{2021_ICML_CLIP} maps images and text into a shared embedding space via contrastive learning so that matching image-text pairs are close together. 
CoOp~\cite{2022_CVPR_Coop} learns continuous context vectors as prompts for CLIP, adapting it to CZSL task with the training set of each dataset. 
DFSP~\cite{2023_CVPR_DFSP} builds soft prompts for attributes and objects separately and fuses them with image features via a cross-modal decomposed fusion module. 
RAPR~\cite{2024_AAAI_RAPR} uses retrieved primitive concepts to enrich representations for better unseen composition recognition. 
Troika~\cite{2024_CVPR_Troika} employs a multi-path prompt architecture plus a cross-modal traction module to jointly model primitives and their compositions. 
PLID~\cite{2024_ECCV_PLID} prompts a large language model to generate a language-informed distribution over class compositions. 
IMAX~\cite{2025_TPAMI_IMAX} embeds attributes and objects in a complex space (with real and imaginary parts) so that phase information captures their dependencies. 
ClusPro~\cite{2025_ICLR_ClusPro} mines multiple clustering-based prototypes for each primitive in embedding space. 
TOMCAT~\cite{2019_NIPS_TOMCAT} overcomes distribution shift of label space caused by unseen compositions through accumulating multimodal knowledge at test time.

\subsection{Main Results}
\label{section4_2:Main_Results}

\textbf{Closed-world Results. }We report the closed-world results of our proposed \model\ and baselines in Table~\ref{tab:cw_results}. 
For UT-Zappos~\cite{2014_CVPR_ut_zappos}, our method surpasses other models by a substantial margin in general. \model\ boosts AUC, HM, and Seen from 48.3\%, 60.2\%, and 74.5\% of our previous version TOMCAT to the new state-of-the-art performance of 52.9\%, 64.3\%, and 78.4\% with 4.6\%, 4.1\%, and 3.9\% improvement, respectively. 
However, \model\ underperforms ClusPro~\cite{2025_ICLR_ClusPro} on the Seen metric because its disentanglement strategy partially distorts the semantics of seen compositions to emulate unseen ones. Although \model\ trails by 2.1\% on Unseen, it surpasses ClusPro~\cite{2025_ICLR_ClusPro} on Seen by a remarkable 7.7\%. 
For the newly proposed benchmark dataset C-Fashion, \model\ achieves 51.5\%, 63.6\%, and 76.6\% on the metrics of AUC, HM, and Seen, providing 3.5\%, 3.9\%, and 2.3\% improvement on our previous TOMCAT. 
For C-GQA~\cite{2021_CVPR_CGE_CGQA}, our method performs significantly better than baselines regarding all metrics. 
Compared to TOMCAT, the new version \model\ increases the four metrics from 16.0\%, 34.0\%, 45.3\%, and 40.1\% to 16.9\%, 35.4\%, 45.6\%, and 41.5\%, achieving 0.9\%, 1.4\%, 0.3\%, and 1.4\% improvement, respectively. 
For the refined MIT-States$^*$, the improved experimental results in AUC (39.4\%) and HM (52.8\%) demonstrate the superiority of our method. 
\model\ achieves state-of-the-art performance in the closed-world setting, demonstrating that our proposed method enables the model to adapt to the distribution shift of label space caused by unseen compositions during testing. 
In addition, \model\ outperforms our previous version TOMCAT, demonstrating that we warm-start the visual priority queue in a principled manner, mitigating the model’s bias toward compositions of previously observed images during testing.

\begin{table*}[t]
    \centering
    \caption{Open-world results on UT-Zappos, C-Fashion, C-GQA, and MIT-States$^*$. In the open-world setting, a post-training feasibility calibration is employed to eliminate implausible compositions from the expanded search space~\cite{2021_CVPR_Compcos_open}.} 
    \resizebox{\linewidth}{!}{
    \begin{tabular}
    {l>{\columncolor{tabcolor}}cccc>{\columncolor{tabcolor}}cccc>{\columncolor{tabcolor}}cccc>{\columncolor{tabcolor}}cccc}
    
        \toprule
         \multirow{2}{*}{Methods} & \multicolumn{4}{c}{UT-Zappos} & \multicolumn{4}{c}{C-Fashion} & \multicolumn{4}{c}{C-GQA} & \multicolumn{4}{c}{MIT-States$^*$}\\
         
         \cmidrule(lr){2-5} \cmidrule(lr){6-9} \cmidrule(lr){10-13} \cmidrule(lr){14-17}
         
          & AUC & HM & Seen & Unseen & AUC & HM & Seen & Unseen & AUC & HM & Seen & Unseen & AUC & HM & Seen & Unseen\\
         \midrule

         CLIP~\cite{2021_ICML_CLIP} \conf{ICML'21} & 2.2 & 11.2 & 15.7 & 20.6 & 6.1 & 18.9 & 26.9 & 14.5 & 0.3 & 4.0 & 7.5 & 4.6  & 3.4 & 13.6 & 35.1 & 12.4 \\ 
         
         CoOp~\cite{2022_CVPR_Coop} \conf{IJCV'22} & 13.2 & 28.9 & 52.1 & 31.5 & 25.4 & 42.2 & 67.5 & 43.0 & 0.7 & 5.5 & 21.0 & 4.6  & 3.2 & 13.0 & 39.3 & 9.7 \\ 
 
         CSP~\cite{2023_ICLR_CSP} \conf{ICLR'23} & 22.7 & 38.9 & 64.1 & 44.1 & 27.7 & 44.3 & 66.0 & 47.0 & 1.2 & 6.9 & 28.7 & 5.2  & 6.2 & 19.2 & 44.4 & 18.5 \\ 
         
         DFSP~\cite{2023_CVPR_DFSP} \conf{CVPR'23} & 30.3 & 44.0 & 66.8 & 60.0 & 30.7 & 47.1 & 69.2 & 50.1 & 2.4 & 10.4 & 38.3 & 7.2 & 9.8 & 25.5 & 46.4 & 26.0 \\ 
           
         Troika~\cite{2024_CVPR_Troika} \conf{CVPR'24} & 33.0 & 47.8 & 66.4 & 61.2 & 33.7 & 48.7 & 68.4 & 56.5 & 2.7 & 10.9 & 40.8 & 7.9 & 11.6 & 27.8 & 49.6 & 28.4 \\ 
         
         PLID~\cite{2024_ECCV_PLID} \conf{ECCV'24} & 30.8 & 46.6 & 67.6 & 55.5 & 28.4 & 45.6 & \textbf{76.7} & 41.3 & 2.5 & 10.6 & 39.1 & 7.5 & 5.9 & 17.6 & 45.1 & 15.3 \\ 

         RAPR~\cite{2024_AAAI_RAPR} \conf{AAAI'24} & 33.3 & 47.9 & 69.4 & 59.4 & 27.3 & 43.8 & 64.9 & 47.3 & \textbf{4.4} & \textbf{14.6} & \textbf{45.5} & \textbf{11.2} & 7.7 & 21.7 & 47.8 & 18.4 \\ 

         IMAX~\cite{2025_TPAMI_IMAX} \conf{TPAMI'25} & 32.3 & 47.5 & 68.4 & 57.3 & 28.6 & 45.6 & 65.3 & 48.1 & 2.6 & 11.2 & 38.7 & 7.9 & 13.9 & 30.8 & 49.7 & 34.2 \\ 

         ClusPro~\cite{2025_ICLR_ClusPro} \conf{ICLR'25} & {39.5} & {54.1} & {71.0} & \textbf{66.2} & 34.7 & 50.2 & 69.3 & 57.0 & {3.0} & {11.6} & {41.6} & {8.3} & 14.2 & \underline{31.2} & 50.2 & 34.0 \\ 

         {TOMCAT}  \conf{NeurIPS'25} & \underline{43.7} & \underline{57.9} & \underline{74.1} & {65.8} & \underline{39.5} & \underline{54.8} & \underline{74.2} & \underline{58.4} & \underline{4.2} & \underline{14.2} & \underline{45.1} & \underline{10.6} & \underline{15.2} & {30.7} & \underline{52.6} & \underline{35.2} \\
         
         \midrule
         
         \textbf{\model\ (Ours)} & \textbf{46.5} & \textbf{59.7} & \textbf{78.3} & \underline{65.9} & \textbf{39.9} & \textbf{55.2} & \underline{74.2} & \textbf{58.9} & {4.0} & {13.7} & {43.3} & {9.7} &\textbf{ 16.2} &\textbf{ 32.2} & \textbf{53.8} & \textbf{35.9} \\
         
        \bottomrule
    \end{tabular}
    }
    
    \label{tab:ow_results}
\end{table*}
\begin{table*}[t]
    \centering
    \caption{Abaltion study of our proposed modules on UT-Zappos, C-Fashion, and C-GQA. AUW is adaptive update weights. PQI for Seen/Unseen Comp. means priority queue initialization for seen/unseen compositions.} 
    \resizebox{1.0\linewidth}{!}{
    \begin{tabular}
    {cccccc>{\columncolor{tabcolor}}cccc>{\columncolor{tabcolor}}cccc>{\columncolor{tabcolor}}cccc}
    
        \toprule
          \multicolumn{6}{c}{Module} & \multicolumn{4}{c}{UT-Zappos} & \multicolumn{4}{c}{C-Fashion} & \multicolumn{4}{c}{C-GQA}\\
         
         \cmidrule(lr){7-10} \cmidrule(lr){11-14} \cmidrule(lr){15-18} 
         
          \makecell{Priority \\ Queue} &\makecell{Textual \\ KAM} & \makecell{Visual \\ KAM} & AUW & \makecell{PQI for \\ Seen Comp.} & \makecell{PQI for \\ Unseen Comp.} & AUC & HM & Seen & Unseen & AUC & HM & Seen & Unseen & AUC & HM & Seen & Unseen\\
         \midrule

          & & & & & & 43.6  & 55.5  & 68.7  & \textbf{74.3}  & 49.6  & 62.3  & 71.5  & \textbf{74.7}  & 15.4  & 33.4  & 44.3  & 39.7 \\
         
         \checkmark & & & & & & 43.3  & 55.7  & 68.6  & 74.1  & 49.3  & 62.1  & 71.5  & 74.4  & 15.4  & 33.2  & 44.5  & 39.7 \\

          & \checkmark & & & & & 45.2  & 58.1  & 74.3  & 68.8  & 51.4  & 63.2  & \textbf{79.0}  & 70.6  & 15.5  & 33.5  & 44.5  & 39.8 \\

         \checkmark & & \checkmark & & & & 43.4  & 55.7  & 68.5  & 74.1  & 49.3  & 62.1  & 71.5  & 74.4  & 14.3  & 31.6  & 44.9  & 36.9\\

         \checkmark & \checkmark & \checkmark &  & & & 46.3  & 58.5  & 76.3  & 68.2  & 51.1  & 62.6  & 78.3  & 71.6  & 15.6  & 33.6  & \textbf{45.6}  & 39.0\\

         \checkmark & \checkmark & \checkmark & \checkmark & & & 48.1  & 59.7  & 74.3  & 72.9  & 51.2  & 63.2  & 76.1  & 73.9  & 15.8  & 33.7  & 45.5  & 39.3 \\
         \checkmark & \checkmark & \checkmark & \checkmark & \checkmark & & 51.1  & 62.1  & \textbf{78.8}  & 72.1  & 51.1  & 63.4  & 74.8  & 73.5  & 15.8  & 34.0  & 45.3  & 39.7 \\
         
         \checkmark & \checkmark & \checkmark & \checkmark & \checkmark & \checkmark & \textbf{52.9}  & \textbf{64.3}  & 78.4  & 73.9  & \textbf{51.5}  & \textbf{63.6}  & 76.6  & 72.5  & \textbf{16.9}  & \textbf{35.4}  & \textbf{45.6}  & \textbf{41.5} \\ 
         
        \bottomrule
    \end{tabular}
    }
    \label{tab:main_ablation}
\end{table*}

\textbf{Open-world Results. }The open-world results are shown in Table~\ref{tab:ow_results}. 
For UT-Zappos~\cite{2014_CVPR_ut_zappos}, our approach achieves notable overall gains over existing models. Compared with the earlier TOMCAT variant, \model\ elevates AUC, HM, and Seen from 43.7\%, 57.9\%, and 74.1\% to 46.5\%, 59.7\%, and 78.3\%, yielding improvement of 2.8\%, 1.8\%, and 4.2\%, respectively. 
For C-Fashion, \model\ attains AUC, HM, and Uneen scores of 39.9\%, 55.2\%, and 58.9\%, corresponding to gains of 0.4\%, 0.4\%, and 0.5\% over the previous TOMCAT results of 39.5\%, 54.8\%, and 58.4\%. 
On MIT-States$^*$, \model\ reaches AUC, HM, Seen, and Unseen scores of 16.2\%, 32.2\%, 53.8\% and 35.9\%. These results mark clear improvements over the earlier TOMCAT, which recorded 15.2\%, 30.7\%, 52.6\%, and 35.2\% on the respective metrics, reflecting gains of 1.0\%, 1.5\%, 1.2\%, and 0.7\%. 
However, \model\ performs worse than RAPR~\cite{2024_AAAI_RAPR} across all metrics. This is primarily because C-GQA dataset involves a large intrinsic compositional search space, and it is further expanded under the open-world setting, which amplifies the difficulty of reliable pseudo-label assignment and exacerbates error accumulation during inference.

\subsection{Ablation Study}
In this section, we conduct ablation studies to assess the contribution of each component in \model\ on UT-Zappos, C-Fashion, and C-GQA in the closed-world setting.

\begin{table*}[t]
    \centering
    \caption{Ablation study of our designed loss on UT-Zappos, C-Fashion, and C-GQA.} 
    {
    \begin{tabular}
    {cc>{\columncolor{tabcolor}}cccc>{\columncolor{tabcolor}}cccc>{\columncolor{tabcolor}}cccc}
    
        \toprule
          \multicolumn{2}{c}{$\mathcal{L}oss$} & \multicolumn{4}{c}{UT-Zappos} & \multicolumn{4}{c}{C-Fashion} & \multicolumn{4}{c}{C-GQA}\\
         
         \cmidrule(lr){3-6} \cmidrule(lr){7-10} \cmidrule(lr){11-14} 
         
          $\mathcal{L}_{PE}$ &$\mathcal{L}_{MCRL}$ & AUC & HM & Seen & Unseen & AUC & HM & Seen & Unseen & AUC & HM & Seen & Unseen \\
         \midrule

          & & 43.6  & 55.5  & 68.7  & \textbf{74.3}  & 49.6  & 62.3  & 71.5  & \textbf{74.7}  & 15.4  & 33.4  & 44.3  & 39.7 \\
         
         \checkmark & & 45.1  & 58.1  & 71.7  & 71.9  & 50.4  & 62.9  & 74.0  & 73.5  & 16.2  & 34.0  & 45.6  & 40.4 \\

          & \checkmark & 36.4  & 52.1  & 57.3  & 70.9  & 43.8  & 57.4  & 68.5  & 70.5  & 11.9  & 28.4  & 35.0  & 40.8 \\

         \checkmark & \checkmark & \textbf{52.9}  & \textbf{64.3}  & \textbf{78.4}  & 73.9  & \textbf{51.5} & \textbf{63.6} & \textbf{76.6} &{ 72.5} & \textbf{16.9}  & \textbf{35.4}  & \textbf{45.6}  & \textbf{41.5}  \\
        \bottomrule
    \end{tabular}
    }
    \label{tab:loss_ablation}
\end{table*}
\begin{table}[t]
    \centering
    \caption{Influence of test order on UT-Zappos, C-Fashion, and C-GQA. SD denotes standard deviation.}
    \resizebox{0.9\linewidth}{!}{
    \begin{tabular}
    {l>{\columncolor{tabcolor}}cc>{\columncolor{tabcolor}}cc>{\columncolor{tabcolor}}cc}
    
        \toprule
         \multirow{2}{*}{Test Order} & \multicolumn{2}{c}{UT-Zappos} & \multicolumn{2}{c}{C-Fashion} & \multicolumn{2}{c}{C-GQA}\\
         
         \cmidrule(lr){2-3} \cmidrule(lr){4-5} \cmidrule(lr){6-7}

         & AUC & HM &AUC & HM & AUC & HM\\
         \midrule
         
        Order 1 & 52.9  & 64.3  & 51.5  & 63.6 & 16.9  & 35.4  \\ 
        Order 2 & 52.0  & 63.7  & 50.9  & 63.6  & 16.5  & 35.2  \\ 
        Order 3 & 51.7  & 63.3  & 51.8  & 63.7  & 16.4  & 35.0 \\
        \midrule
        Mean & 52.2 & 63.8 & 51.4 & 63.6 & 16.6 & 35.2 \\
        SD & 0.6 & 0.5 & 0.5 & 0.1 & 0.3 & 0.2 \\
        \bottomrule
    \end{tabular}
    }
    \label{tab:test_order}
\end{table}
\begin{table}[t]
    \centering
    \vspace{-1.5pt}
    \caption{Ablation Study of prediction probability during inference on UT-Zappos, C-Fashion, and C-GQA. 
    }
    \resizebox{\linewidth}{!}{
    \begin{tabular}
    {l>{\columncolor{tabcolor}}cccc>{\columncolor{tabcolor}}cccc>{\columncolor{tabcolor}}cccc}
    
        \toprule
         \multirow{2}{*}{Prediction Probability} & \multicolumn{2}{c}{UT-Zappos} & \multicolumn{2}{c}{C-Fashion} & \multicolumn{2}{c}{C-GQA} \\
         
         \cmidrule(lr){2-3} \cmidrule(lr){4-5} \cmidrule(lr){6-7}
         
          & AUC & HM & AUC & HM & AUC & HM \\
         \midrule
         
         Text-Only & 52.5  & 64.1  & 50.9  & 63.5  & 16.8  & 35.3
 \\ 
        Visual-Only & 12.1  & 27.1  & 9.1  & 25.9  & 2.4  & 13.7
 \\ 
        Multimodal & \textbf{52.9}  & \textbf{64.3}  & \textbf{51.5}  & \textbf{63.6}  & \textbf{16.9}  & \textbf{35.4}
 \\ 
         
        \bottomrule
    \end{tabular}
    }
    
    \label{tab:prediction_probability}
\end{table}

\textbf{Effectiveness of Main Modules.} As the most significant contribution of our work, we first accumulate visual and textual knowledge at test time for CZSL with the proposed modules in Table~\ref{tab:main_ablation}. Specifically, compared with the base model, incorporating the priority queue and the subsequent visual KAM results in performance degradation across the three datasets. This is because the visual prototypes are updated by pseudo-labeled samples and visual KAM with insufficient semantic guidance of original textual prototypes, which can introduce noisy and biased visual update. In other words, in the absence of textual KAM and adaptive update weights (AUW), visually driven accumulation may cause prototype drift, especially under label-space shift, leading to suboptimal adaptation. 
Adding only the textual KAM allows refinement of the textual prototypes under the entropy minimization objective, making them more confident and discriminative, which helps the model better distinguish compositions. 
When both visual and textual KAMs are applied, the model benefits from complementary multimodal refinement. Visual KAM continuously calibrates the distribution of visual prototypes based on incoming test samples, while textual KAM captures the evolving semantic structure of the label space, leading to the most accurate and robust prototypes and achieving the best overall performance. 
The improvement brought by AUW can be attributed to determining the extent of prototype update based on the similarity, avoiding indiscriminately assigning equal treatment to all compositions regardless of their familiarity.
In this version, we leverage training images and the mapping between seen and unseen textual prototypes to initialize visual prototypes of seen and unseen compositions, respectively. 
Initializing only for the seen compositions helps mitigate the model’s bias toward previously observed compositions, leading to the performance improvement. However, this setting still causes the model to favor predictions on seen compositions. When prototypes of unseen compositions are also initialized, the model achieves the optimal performance. 

\textbf{Effectiveness of Each Loss. }
As reported in Table~\ref{tab:loss_ablation}, incorporating $\mathcal{L}_{PE}$ into the base model yields a clear performance gain, demonstrating that the prediction entropy loss effectively guides the model to adapt to the shifted label distribution by encouraging the model to produce confident predictions with less uncertainty. In contrast, adopting only the constraint term $\mathcal{L}_{MCRL}$ tends to indiscriminately enforce the alignment of multimodal prototypes by merely minimizing their distances, while neglecting discriminative structure and distributional uncertainty. This overly rigid regularization can introduce suboptimal prototype configurations and consequently impair overall performance. By jointly optimizing $\mathcal{L}_{PE}$ and $\mathcal{L}_{MCRL}$, \model\ strikes a favorable balance between adaptive uncertainty minimization and structured prototype alignment, leading to the best results among all variants and thereby validating the effectiveness of our algorithmic design.

\begin{table}[t]
    \centering
    \caption{Influence of initialization strategies of KAMs on UT-Zappos, C-Fashion and C-GQA.}
    \resizebox{0.955\linewidth}{!}{
    \begin{tabular}
    {l>{\columncolor{tabcolor}}cc>{\columncolor{tabcolor}}cc>{\columncolor{tabcolor}}cc}
    
        \toprule
         \multirow{2}{*}{Initialization} & \multicolumn{2}{c}{UT-Zappos} & \multicolumn{2}{c}{C-Fashion} & \multicolumn{2}{c}{C-GQA}\\
         
         \cmidrule(lr){2-3} \cmidrule(lr){4-5} \cmidrule(lr){6-7}

         & AUC & HM &AUC & HM & AUC & HM\\
         \midrule
         
        Uniform Random & 43.5  & 56.1  & 49.7  & 62.7  & 15.6  & 34.0 \\
        Normal Random & 51.1  & 62.3  & 49.2  & 62.7  & 16.3  & 35.0  \\
        Random Walking & 51.2  & 63.3  & 50.6  & 63.3  & \textbf{16.9}  & 35.3  \\
        All Zeros & \textbf{52.9}  & \textbf{64.3}  & \textbf{51.5} & \textbf{63.6} & \textbf{16.9}  & \textbf{35.4}  \\ 
        
        \bottomrule
    \end{tabular}
    }
    \label{tab:KAM_initialization}
\end{table}
\begin{table}[t]
    \centering
    \caption{Influence of different test-time methods based on base model on UT-Zappos, C-Fashion, and C-GQA.}
    \resizebox{0.955\linewidth}{!}{
    \begin{tabular}
    {l>{\columncolor{tabcolor}}cc>{\columncolor{tabcolor}}cc>{\columncolor{tabcolor}}cc}
    
        \toprule
         \multirow{2}{*}{Method} & \multicolumn{2}{c}{UT-Zappos} & \multicolumn{2}{c}{C-Fashion} & \multicolumn{2}{c}{C-GQA}\\
         
         \cmidrule(lr){2-3} \cmidrule(lr){4-5} \cmidrule(lr){6-7}

         & AUC & HM &AUC & HM & AUC & HM\\
         \midrule
         
        Online-TPS~\cite{2025_WACV_TPS} & 44.6  & 57.3  & 49.3  & 61.4  & 14.8  & 32.7  \\ 
        TDA~\cite{2024_CVPR_TDA} & 41.6  & 54.6  & 47.7  & 60.3  & 14.4  & 32.0  \\ 
        DPE~\cite{2024_NIPS_DPE} & 43.3  & 56.0  & 49.3  & 62.1  & 15.3  & 33.5 \\ 

        \midrule
        \textbf{\model\ (Ours)} & \textbf{52.9}  & \textbf{64.3}  & \textbf{51.5} & \textbf{63.6} & \textbf{16.9}  & \textbf{35.4}  \\ 
        \bottomrule
    \end{tabular}
    }
    \label{tab:tta_ablation}
\end{table}

\textbf{Influence of Test Order. }
As shown in Table~\ref{tab:test_order}, \model\ exhibits strong robustness to different test sample orders. The performance variations across three random orders remain minimal on all three datasets, confirming that the proposed test-time knowledge accumulation pipeline is stable and suffers from little sensitivity to data sequencing. This further validates the effectiveness of the priority-based memory mechanism, together with reasonable warm-start strategy, in maintaining consistent adaptation behavior under dynamic test conditions.

\begin{table*}[t]
    \centering

    \begin{minipage}{0.48\linewidth}
        \centering
        \caption{Ablation study of fine-tuning CLIP on the training set of UT-Zappos, C-Fashion, and C-GQA, respectively.}
    {
    \begin{tabular}
    {lc>{\columncolor{tabcolor}}cccc}
        \toprule
         Dataset & \makecell{training\\ tuning} & AUC & HM & Seen & Unseen \\
         \midrule

        \multirow{2}{*}{UT-Zappos} & \ding{55} & 3.3 & 11.6 & 10.6 & 52.3 \\
         & \checkmark & \textbf{52.9} & \textbf{64.3} & \textbf{78.4} & \textbf{73.9} \\
         \midrule

        \multirow{2}{*}{C-Fashion} & \ding{55} & 11.1 & 25.3 & 26.0 & 61.1 \\
        & \checkmark & \textbf{51.5} & \textbf{63.6} & \textbf{76.6} & \textbf{72.5} \\
        \midrule

        \multirow{2}{*}{C-GQA} & \ding{55} & 1.7 & 8.9 & 7.9 & 25.9 \\
        & \checkmark & \textbf{16.9} & \textbf{35.4} & \textbf{45.6} & \textbf{41.5} 
 \\
         
        \bottomrule
    \end{tabular}
    }
    
    \label{tab:clip_tuning}
    \end{minipage}
    \hfill
    \begin{minipage}{0.48\linewidth}
        \centering
        \caption{Ablation study of test-time tuning of CLIP~\cite{2021_ICML_CLIP}, CSP~\cite{2023_ICLR_CSP}, and the base model on Ut-Zappos and C-GQA.}
    {
    \begin{tabular}
    {lc>{\columncolor{tabcolor}}cc>{\columncolor{tabcolor}}cc}
    
        \toprule
         \multirow{2}{*}{Method} & \multirow{2}{*}{\makecell{test-time\\ tuning}} & \multicolumn{2}{c}{UT-Zappos}  & \multicolumn{2}{c}{C-GQA}\\
         \cmidrule(lr){3-4} \cmidrule(lr){5-6}

         & & AUC & HM &AUC & HM\\
    \midrule
    \multirow{2}{*}{CLIP~\cite{2021_ICML_CLIP}} 
    & \ding{55} & \textbf{5.0} & \textbf{15.6} & 1.4 & 8.6  \\
    & \checkmark & 3.3 & 11.6 & \textbf{1.7} & \textbf{8.9} \\
    \midrule
    \multirow{2}{*}{CSP~\cite{2023_ICLR_CSP}} 
    & \ding{55} & 33.0 & 46.6 & 6.2 & 20.5 \\
    & \checkmark & \textbf{40.3} & \textbf{55.4} & \textbf{7.3} & \textbf{22.8} \\
    \midrule
    \multirow{2}{*}{Base Model} 
    & \ding{55} & 43.6 & 55.5 & 15.4 & 33.4 \\
    & \checkmark & \textbf{52.9} & \textbf{64.3} & \textbf{16.9} & \textbf{35.4}\\
    \bottomrule
    \end{tabular}
    }
    \label{tab:ablation_test}
    \end{minipage}

\end{table*}

\begin{figure*}[t]
\centering
\begin{minipage}{\textwidth}
\centering
\subfloat[UT-Zappos\label{fig:alpha_ut_zappos}]{
    \includegraphics[width=0.31\textwidth]{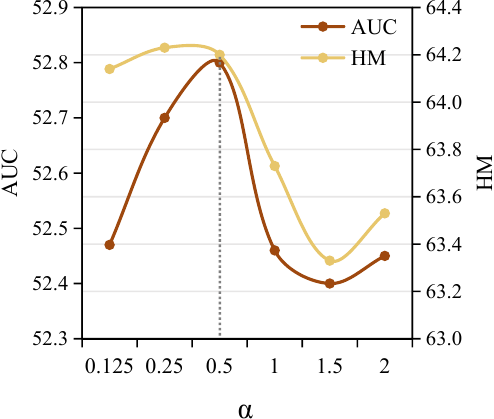}
}\hfill
\subfloat[C-Fashion\label{fig:alpha_mit_states}]{
    \includegraphics[width=0.31\textwidth]{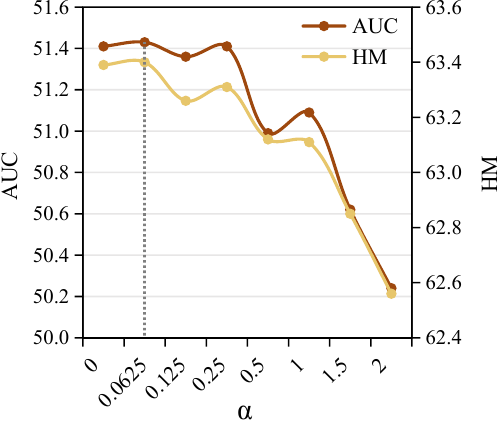}
}\hfill
\subfloat[C-GQA\label{fig:alpha_cgqa}]{
    \includegraphics[width=0.31\textwidth]{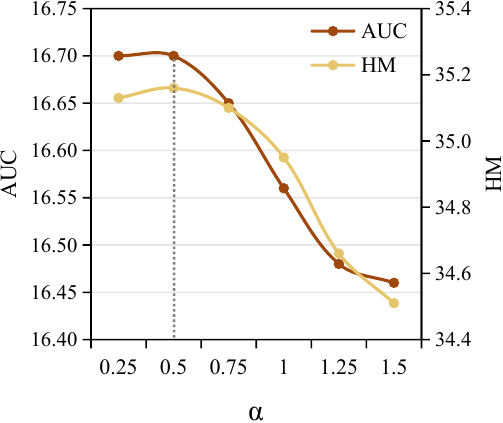}
}
\end{minipage}

\caption{Influence of multimodal balance factor $\alpha$ on UT-Zappos, C-Fashion, and C-GQA.}
\label{fig:alpha}
\end{figure*}

\begin{figure*}[t]
\centering
\begin{minipage}{\textwidth}
\centering
\subfloat[UT-Zappos\label{fig:theta_ut_zappos}]{
    \includegraphics[width=0.31\textwidth]{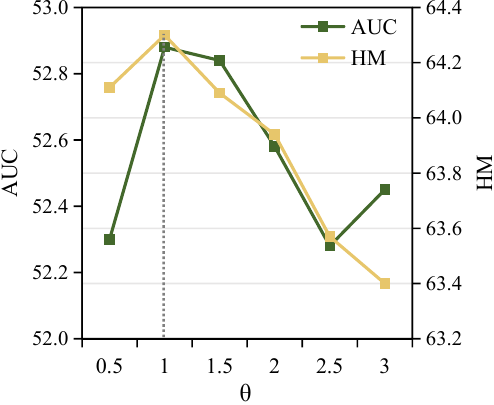}
}\hfill
\subfloat[C-Fashion\label{fig:theta_mit_states}]{
    \includegraphics[width=0.31\textwidth]{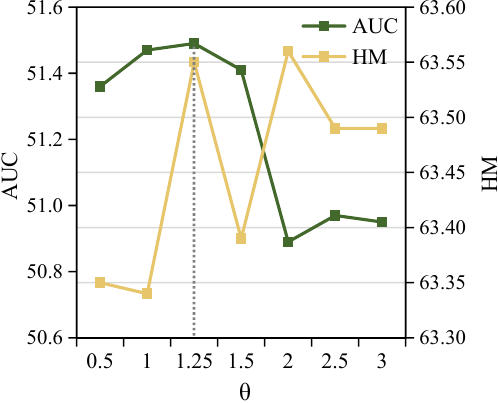}
}\hfill
\subfloat[C-GQA\label{fig:theta_cgqa}]{
    \includegraphics[width=0.31\textwidth]{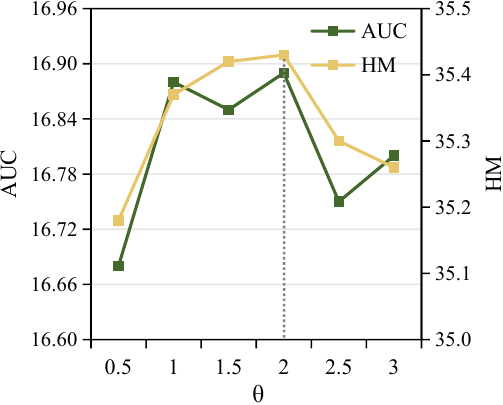}
}
\end{minipage}

\caption{Influence of the update control factor in AUW $\theta$ on UT-Zappos, C-Fashion, and C-GQA.}
\label{fig:theta}
\end{figure*}

\textbf{Effectiveness of Prediction Probability During Inference. }
Table~\ref{tab:prediction_probability} shows the influence of the prototype source used to derive prediction probabilities during inference. Using textual-only or visual-only prototypes leads to clear performance degradation across all datasets, indicating that relying on a single modality provides incomplete or biased semantic supervision. Specifically, textual prototypes lack instance-level visual details, while visual prototypes are susceptible to noise accumulated from pseudo-labels and feature domain shift. In contrast, incorporating both textual and visual prototypes consistently achieves the best performance, demonstrating that both textual and visual prototypes provide complementary cues for more robust and accurate predictions.

\textbf{Influence of Initialization Strategies of KAMs. }
Table~\ref{tab:KAM_initialization} investigates the influence of different KAM initialization strategies. Random initializations, including uniform, normal, and random-walk schemes, tend to introduce substantial noise at the beginning of test, which disrupts the originally well-structured prototypes learned during training and can cause partial damage of previously acquired knowledge. In contrast, initializing KAM with all zeros avoids injecting artificial perturbations at the test onset, allowing the model to preserve its trained representations while gradually refining prototypes through reliable test-time evidence, leading to more stable and superior results.

\begin{table*}[t]
    \centering
    \caption{Long-tailed results on C-Fashion and MIT-States$^*$.}
    \resizebox{\linewidth}{!}{
    \begin{tabular}
    {lccc>{\columncolor{tabcolor}}c>{\columncolor{tabcolor}}cccc>{\columncolor{tabcolor}}c>{\columncolor{tabcolor}}c}
    
        \toprule
         \multirow{2}{*}{Methods} & \multicolumn{5}{c}{C-Fashion} & \multicolumn{5}{c}{MIT-States$^*$}\\
         
         \cmidrule(lr){2-6} \cmidrule(lr){7-11} 
         
          & Head Acc$\uparrow$ & Body Acc$\uparrow$ & Tail Acc$\uparrow$ & All Acc$\uparrow$ &All Std$\downarrow$ & Head Acc$\uparrow$ & Body Acc$\uparrow$ & Tail Acc$\uparrow$ & All Acc$\uparrow$ & All Std$\downarrow$ \\
         \midrule

         
         CLIP~\cite{2021_ICML_CLIP} \conf{ICML'21} &  24.7  & 25.5  & 23.9  & 24.8  & \underline{26.1} & 30.4  & 36.8  & 40.1  & 35.9  & 35.9  \\  
         
         CoOp~\cite{2022_CVPR_Coop} \conf{IJCV'22} & 35.2  & 27.6  & 27.4  & 29.8  & 31.8 & 33.4  & 41.8  & 42.4  & 39.5  & 36.5 \\ 
         
         
         CSP~\cite{2023_ICLR_CSP} \conf{ICLR'23} & 34.2  & 25.3  & 21.7  & 26.9  & 30.2 & 39.5  & 38.1  & 32.8  & 36.9  & 34.3 \\ 
         
         
         
         DFSP~\cite{2023_CVPR_DFSP} \conf{CVPR'23} &  35.3  & 27.0  & 27.9  & 29.7  & 31.3 & 39.5  & 38.5  & 37.1  & 38.4  & 34.2 \\ 
         
         
         Troika~\cite{2024_CVPR_Troika} \conf{CVPR'24} & 36.1  & 30.0  & 35.8  & 33.6  & 33.1 & 44.8  & \underline{43.9}  & 41.7  & 43.5  & 34.8  \\ 
         
         PLID~\cite{2024_ECCV_PLID} \conf{ECCV'24} & 29.0  & 28.5  & 25.8  & 27.8  & 31.3 & 42.7  & 32.5  & 28.6  & 34.3  & \underline{32.2}  \\ 

         RAPR~\cite{2024_AAAI_RAPR} \conf{AAAI'24} & 27.7  & 25.9  & 23.7  & 25.8  & 26.9 & 42.4  & 34.6  & 34.6  & 37.0  & 33.5 \\ 

         IMAX~\cite{2025_TPAMI_IMAX} \conf{TPAMI'25} & 34.9  & 30.7  & 31.7  & 32.3  & 33.2 & 43.0  & 39.6  & 39.9  & 40.7  & 33.0 \\ 

         ClusPro~\cite{2025_ICLR_ClusPro} \conf{ICLR'25} &38.4  & 33.6  & \underline{37.3}  & 36.2  & 31.7  & 42.7  & 40.1  & 39.1  & 50.5  & 32.8 \\ 

        TOMCAT~\cite{2019_NIPS_TOMCAT} \conf{NeurIPS'25} & \underline{39.9} & \underline{33.8}  & 36.6  & \underline{36.5}  & 33.8  & \underline{47.1}  & 42.0  & 42.8  & \underline{43.8}  & 35.0 \\ 
        
         \midrule
         
         \textbf{\model} & \textbf{40.1}  & \textbf{38.7}  & \textbf{40.4}  & \textbf{39.6}  & \textbf{25.3}  & \textbf{48.1}  & \textbf{44.7}  & \textbf{47.9}  & \textbf{46.2}  & \textbf{31.0} \\ 
         
        \bottomrule
    \end{tabular}
    }
    
    \label{tab:long_tail_results}
\end{table*}
\begin{figure*}[t]
    \centering
    \subfloat[UT-Zappos\label{fig:ut_zappos_trenda}]{\includegraphics[width=0.32\textwidth]
    {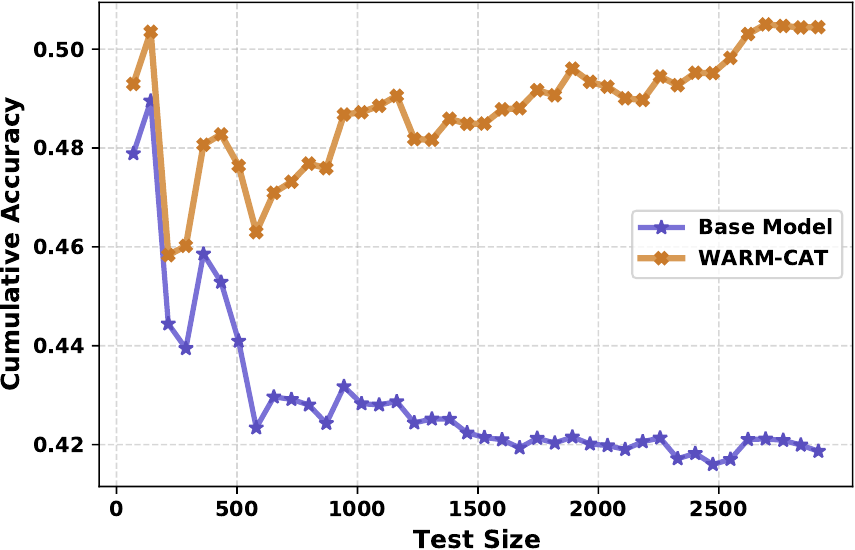}}
    \subfloat[C-Fashion\label{fig:c_fashion_trenda}]{\includegraphics[width=0.32\textwidth]
    {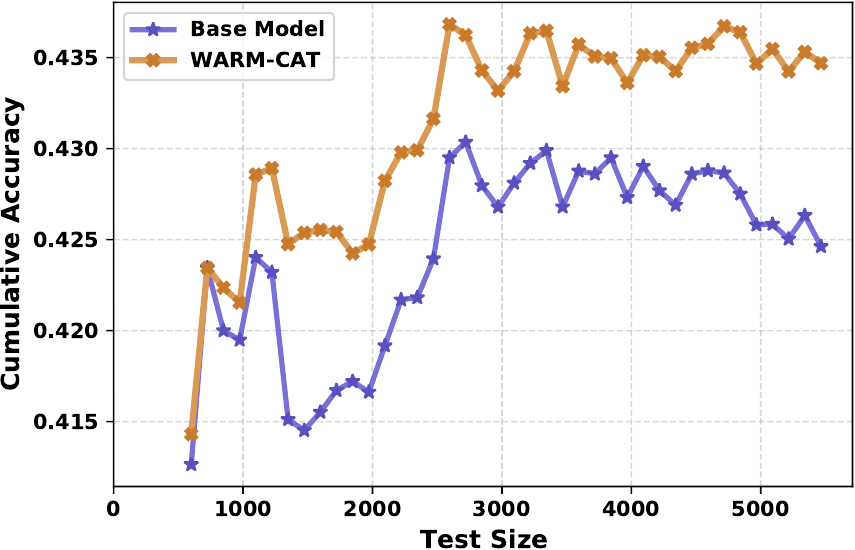}}
    \subfloat[C-GQA\label{fig:cgqa_trenda}]{\includegraphics[width=0.32\textwidth]{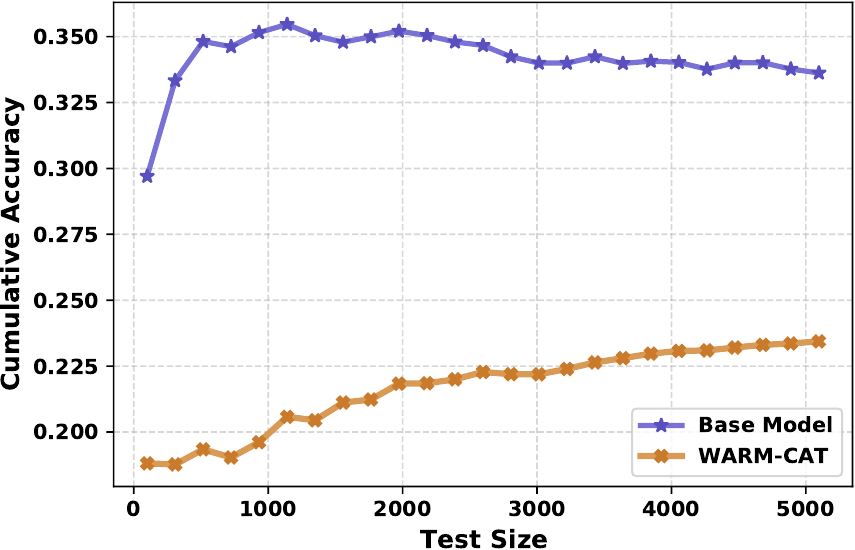} } 
    \caption{Cumulative classification accuracy with increasing test sample size on three datasets.}
    \label{fig:trend}
\end{figure*}

\textbf{Effectiveness of Tuning During Training and Test.} As shown in Table~\ref{tab:clip_tuning}, we ablate the trainable prompts and visual adapters during the training phase on the three datasets. Consistent improvements are observed across all datasets when adopting the fine-tuned CLIP, which indicates that pre-training alone does not sufficiently endow CLIP with compositional reasoning capability. In other words, adapting the representation learned by CLIP to the CZSL task with training set is critical for equipping the model with attribute–object compatibility and enabling more discriminative compositional cues. 
In addition, we conduct the ablation study of test-time tuning framework based on CLIP~\cite{2021_ICML_CLIP}, CSP~\cite{2023_ICLR_CSP}, and the base model in Table~\ref{tab:ablation_test}, and compare different test-time methods in Table~\ref{tab:tta_ablation}. 
Since CLIP has not been tuned on the CZSL training set, its initial prototypes lack reliable compositional priors, causing test-time accumulation to reinforce noisy pseudo-labels and amplify erroneous knowledge, thereby leading to performance degradation. In contrast, both CSP and the base model are trained with explicit compositional data, yielding more accurate and structure-aware prototypes when equipped with test-time \model\ compared with their non-test-time-tuning counterparts. Moreover, \model\ outperforms other test-time methods, which shows knowledge accumulation continuously refines class representations with informative test samples, enabling consistent performance gains.


\textbf{Impact of Multimodal Balance Factor $\alpha$.} The hyper-parameter $\alpha$ controls the relative contributions of visual and textual predictions during inference. As shown in Fig.~\ref{fig:alpha}, the performance exhibits a non-monotonic trend with respect to $\alpha$, peaking at an intermediate value. When $\alpha$ is small, the fused predictions rely excessively on the textual modality and focus less on visual evidence, which constrains discriminability for visually ambiguous compositions and limits the complementary exploitation of cross-modal cues. Conversely, a large $\alpha$ overemphasizes visual cues and weakens semantic guidance from textual prototypes, making predictions more vulnerable to visual noise and prototype misalignment. Proper calibration of $\alpha$ enables a balanced integration of multimodal information, thereby facilitating robust compositional inference.

\textbf{Impact of Update Control Factor in AUW $\theta$.} It is observed in Fig.~\ref{fig:theta} that as the value of $\theta$ increases, the performance first improves and then degrades in general. This behavior indicates a trade-off between stability and plasticity in prototype update. When $\theta$ is small, the update strength is overly conservative, leading to insufficient update of the prototypes to test images that exhibit large appearance variations, thereby limiting the model’s ability to effectively absorb informative knowledge of new distribution. In contrast, an excessively large $\theta$ induces aggressive prototype update, which makes the accumulation vulnerable to noisy pseudo-labels and causes over-adaptation, causing performance deterioration.


\subsection{Qualitative Analysis}

In this section, we conduct a qualitative analysis of \model\ 
on the four datasets 
in the closed-world setting.

\textbf{Long-Tailed Distribution Analysis in CZSL.} The long-tailed distribution phenomenon is a fundamental challenge in classification tasks~\cite{2022_IJCV_longtail2,2025_arxiv_longtail1}, and this holds equally in compositional recognition~\cite{2022_CVPR_comp_longtail,2024_AAAI_Revealing}. While previous works analyze the existence of a long-tailed prior among compositional classes in CZSL~\cite{2024_AAAI_Revealing}, it does not propose explicit class-level metrics for quantifying long-tailed effects nor report corresponding empirical analysis. In our work, we address this gap by ordering all test classes by their image counts and partitioning them into three disjoint \textbf{Head}, \textbf{Body}, and \textbf{Tail} groups in a 3 : 4 : 3 ratio. We report per-group class-averaged accuracy(\textbf{Acc}) of the three groups and \textbf{All} classes, as well as the Standard Deviation (\textbf{SD}) of \textbf{All} classes of our method and the baselines. This stratified reporting sheds light on how recognition performance varies across the distributional spectrum from frequent to rare compositional classes.

\begin{figure*}[t]
    \centering
    \includegraphics[width=0.98\linewidth]{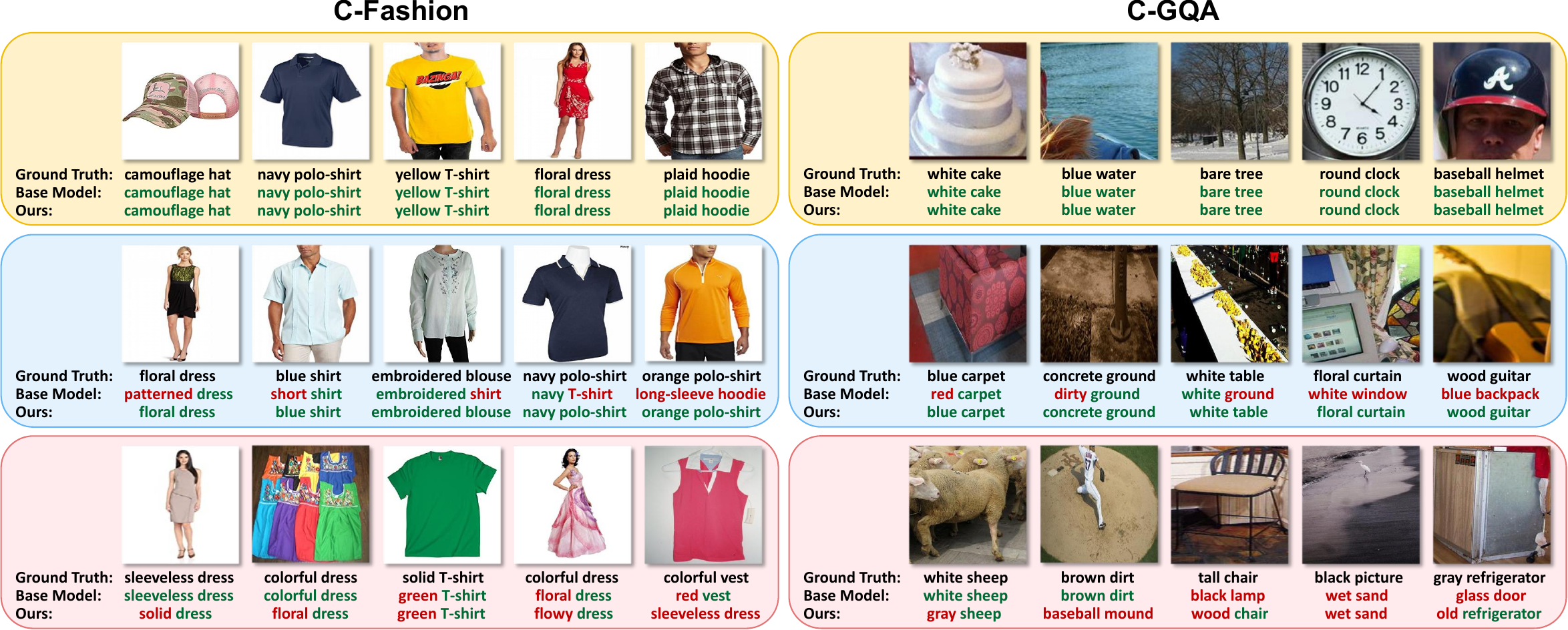} 
    \caption{Successful and failure cases on C-Fashion and C-GQA. The predictions of images from \model\ (Ours) with the base model are reported. The successful and failure results are marked in \textcolor[HTML]{006633}{green} and \textcolor[HTML]{990000}{red}, respectively.}
    \label{fig:cases}
\end{figure*}
\begin{figure*}[t]
    \centering
    \subfloat[Test starting point.]{%
        \includegraphics[width=0.38\textwidth]{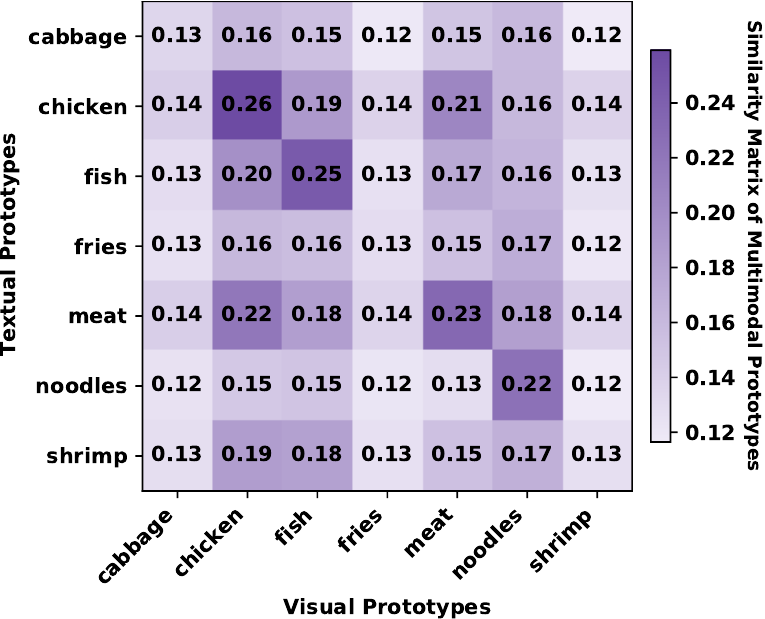}
        \label{fig:sim_mat_start}
    }
    \hspace*{0.07\textwidth} 
    \subfloat[Test end point.]{%
        \includegraphics[width=0.38\textwidth]{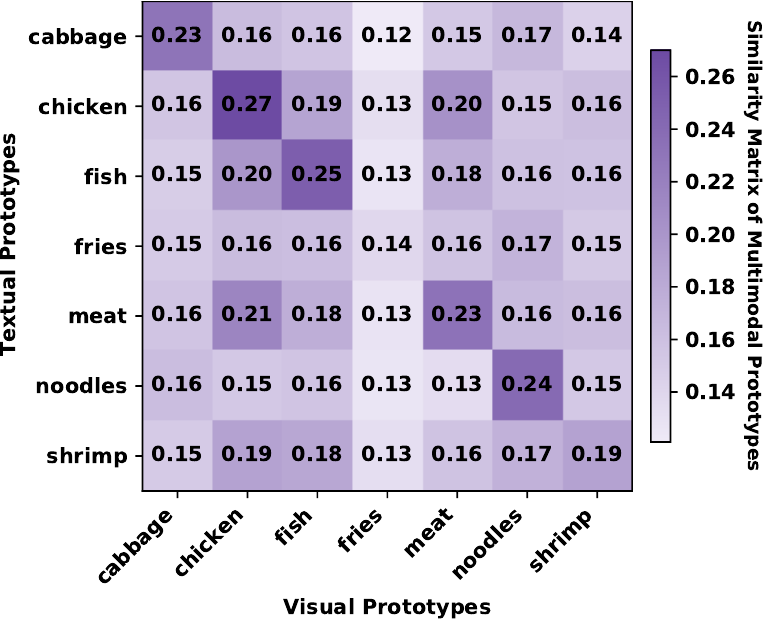}
        \label{fig:sim_mat_end}
    }
    \caption{Comparison of multimodal prototype similarity at the test onset and endpoint on C-GQA. All unseen compositions consisting of the attribute \texttt{cooked} and its corresponding objects (\eg, \texttt{noodles}, \texttt{shrimp}...) are selected.}
    \label{fig:sim_mat}
\end{figure*}

As shown in Table~\ref{tab:long_tail_results}, we experimentally evaluate the mainstream methods and analyze their performance on C-Fashion and MIT-States$^*$ under long-tailed distributions. Existing approaches demonstrate varying degrees of sensitivity to the long-tailed distribution, generally favoring head compositions while underperforming on rare tail classes. In comparison, \model\ achieves more balanced performance across head, body, and tail compositions, resulting in both higher overall accuracy and reduced class-wise deviation. 
Specifically, on C-Fashion, \model\ boosts TOMCAT in tail accuracy from 36.6\% to 40.4\% (+3.8\%), raises overall accuracy from 36.5\% to 39.6\% (+3.1\%), and substantially reduces performance deviation from 33.8\% to 25.3\% (–8.5\%). Similarly, on MIT-States$^*$, \model\ increases tail accuracy from 42.8\% to 47.9\% (+5.1\%), improves all accuracy from 43.8\% to 46.2\% (+2.4\%), and lowers the all standard deviation from 35.0\% to 31.0\% (–4.0\%).
This suggests that our method provides a more stable and equitable recognition capability across the full long-tailed compositional spectrum.

This behavior can be attributed to the fact that all existing models, including TOMCAT~\cite{2019_NIPS_TOMCAT}, rely solely on textual prototypes as the only matching source at the test onset. Since these textual prototypes are learned exclusively from seen compositions during training, their semantic representations are inherently biased toward frequent head categories. Although TOMCAT incrementally enriches visual prototypes by accumulating knowledge from test images, the long-tailed distribution causes most test samples to come from head classes, leading to the lack of visual cues of rare tail categories and, consequently, poorly constructed visual representations in the tail regime. In contrast, \model\ warm-starts visual prototypes for all classes by leveraging both seen compositions and the learned mapping relationships, providing a more complete and balanced starting point for prototype adaptation. This design effectively alleviates the issue of tail prototype deficiency and enables more stable knowledge acquisition across long-tail categories, thereby accounting for the substantial improvements observed on tail classes and the reduced variance reported in Table~\ref{tab:long_tail_results}.

\textbf{Cumulative Classification Accuracy Analysis.}
Fig.~\ref{fig:trend} illustrates how cumulative accuracy evolves throughout the test stream. At the beginning, \model\ outperforms the base model due to the effective warm-start strategy of the visual priority queue, which provides stronger visual information before test starts. As testing progresses, the performance trajectories of the two models diverge markedly. Specifically, the base model performs worse than \model\ under the label distribution shift. The widening gap indicates that \model\ continuously absorbs useful information from the test stream and incrementally refines its prototypes, leading to progressively improved predictions.

\textbf{Case Study.}
In Fig.~\ref{fig:cases}, we report the successful and failure predictions of images from our \model\ and the base model on C-Fashion and C-GQA. 
It is observed that both models perform well on simple test images, such as a \texttt{yellow T-shirt} or a \texttt{white cake}. However, \model\ demonstrates superior performance on more challenging images that require reasoning about attribute prominence. For instance, in an image of an \texttt{orange polo-shirt}, the base model predicts the attribute as \texttt{long-sleeve}, whereas the more salient attribute is clearly the color \texttt{orange}. Similarly, for an image of a \texttt{white table}, the base model confuses the \texttt{brown ground} with the \texttt{white table}, while \model\ gives the correct prediction. This improvement can be attributed to \model’s test‑time comprehensive knowledge accumulation mechanism, which dynamically updates multimodal prototypes by aggregating high‑confidence visual evidence and semantic information. Nevertheless, 
some failure cases. For an image of a \texttt{colorful dress}, \model\ predicts the attribute \texttt{floral}, which is present but not the dominant attribute. Additionally, due to label noise, the label of the image of a \texttt{white sheep} should be \texttt{gray}, but our model still predicts correctly.

\textbf{Analysis of Multimodal Prototype Evolution. }
Figure~\ref{fig:sim_mat} illustrates the pairwise cosine similarities between multimodal prototypes for randomly selected compositions that include the attribute \texttt{cooked} alongside their associated objects. From the beginning to the end of testing, the diagonal entries in the similarity matrix become noticeably larger, indicating that \model\ progressively strengthens the alignment between visual and textual modalities. At the same time, off-diagonal elements also increase, which can be attributed to the fact that all \texttt{cooked} food compositions naturally form a semantically coherent cluster. The higher similarities among these off-diagonal pairs demonstrate that \model\ effectively captures and preserves the shared semantic structure of related compositions, enabling better discrimination ability. 

\section{Conclusion}
\label{section5:Conclusion}
In this work, we consider the issue of label distribution shift in CZSL, which arises from novel compositions recomposed from attributes and objects. Therefore, we propose \model\ to accumulate comprehensive knowledge of visual and textual modalities from unsupervised data at test time to overcome the challenge. Specifically, we update textual and visual prototypes by multimodal knowledge accumulation modules, and use adaptive update weights to control the degree. 
Meanwhile, a priority queue is leveraged that stores historical images to obtain the visual knowledge. 
We warm-start it with images of seen compositions and generate virtual visual prototypes for unseen compositions by leveraging the learned mapping from textual prototypes, thereby mitigating the model’s bias toward historically observed compositions. 
To enable a more comprehensive evaluation of CZSL, we introduce the C-Fashion dataset and refine the MIT-States dataset.
Extensive experiments, including results under long-tailed distribution, demonstrate the superiority of our \model\ in both closed-world and open-world settings.

\section*{Acknowledgments}
This work was supported by the Fundamental Research Funds for the Central Universities (No. 2025JBZX059), the Beijing Natural Science Foundation (No.4242046) and the Natural Science Foundation of Hebei Province (No. F2025105018).

{
    \bibliographystyle{IEEEtran}
    \bibliography{IEEE}
}

\begin{IEEEbiography}[{\includegraphics[width=1in,height=1.25in,clip,keepaspectratio]{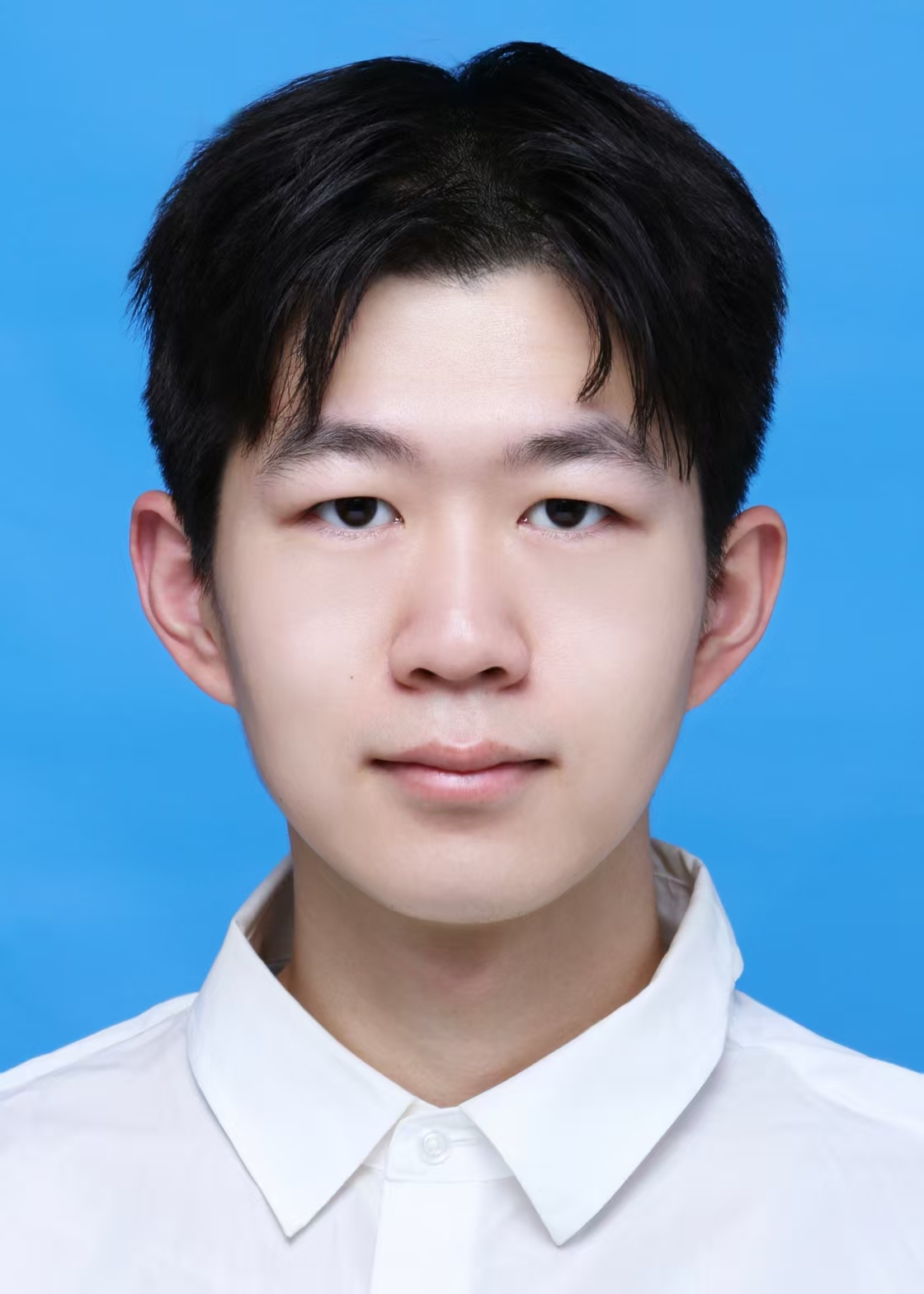}}]{Xudong Yan}
received the B.S. degree from Qingdao University, China, in 2023. He is currently pursuing the Ph.D. degree at the School of Computer Science and Technology, Beijing Jiaotong University. His research interests include applying vision-language models (VLMs) to real-world scenarios, particularly in zero-shot learning and test-time adaptation. He has published papers in \textit{IJCAI} and \textit{NeurIPS}.
\end{IEEEbiography}

\begin{IEEEbiography}[{\includegraphics[width=1in,height=1.25in,clip,keepaspectratio]{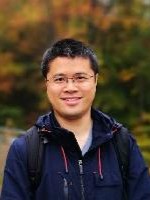}}]{Songhe Feng}
received the B.S. and Ph.D degrees from the School of Computer and Information Technology, Beijing Jiaotong University, Beijing, China, in 2003 and 2009, respectively. He is currently a full professor with the School of Computer Science and Technology, Beijing Jiaotong University. His research interests are machine learning and its applications, such as multi-view clustering, weakly-supervised multi-label learning, zero-shot learning and domain adaptation. He has published more than 100 peer-reviewed papers, including those in highly regarded journals and conferences such as the \textit{IEEE Transactions on Pattern Analysis and Machine Intelligence, IEEE Transactions on Knowledge and Data Engineering, IEEE Transactions on Image Processing, IEEE Transactions on Cybernetics, IEEE Transactions on Multimedia, IEEE Transactions on Circuits and Systems for Video Technology, ACM Transactions on Knowledge Discovery from Data, ICML, NeurIPS, CVPR, ICCV, AAAI, IJCAI, ACM Multimedia}, etc. He has been a visiting scholar with Concordia University, Canada and Michigan State University, USA in 2024 and 2014, respectively.
\end{IEEEbiography}

\begin{IEEEbiography}[{\includegraphics[width=1in,height=1.25in,clip,keepaspectratio]{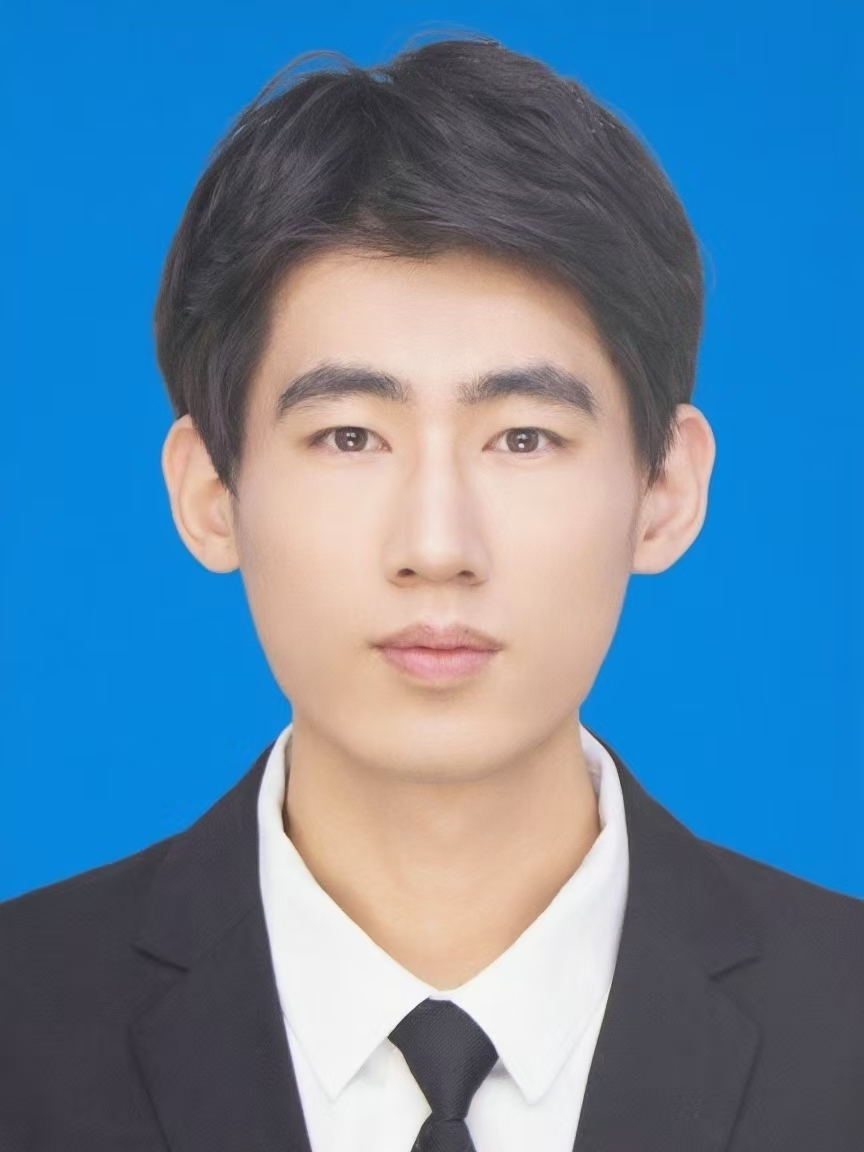}}]{Jiaxin Wang}
received the B.S. degree in Computer Science and Technology from Nanjing Tech University, Nanjing, China. He is currently pursuing the M.S. degree in Computer Science and Technology at Beijing Jiaotong University, Beijing, China. His current research interests include machine learning and computer vision, with a particular focus on compositional zero-shot learning.
\end{IEEEbiography}

\begin{IEEEbiography}[{\includegraphics[width=1in,height=1.25in,clip,keepaspectratio]{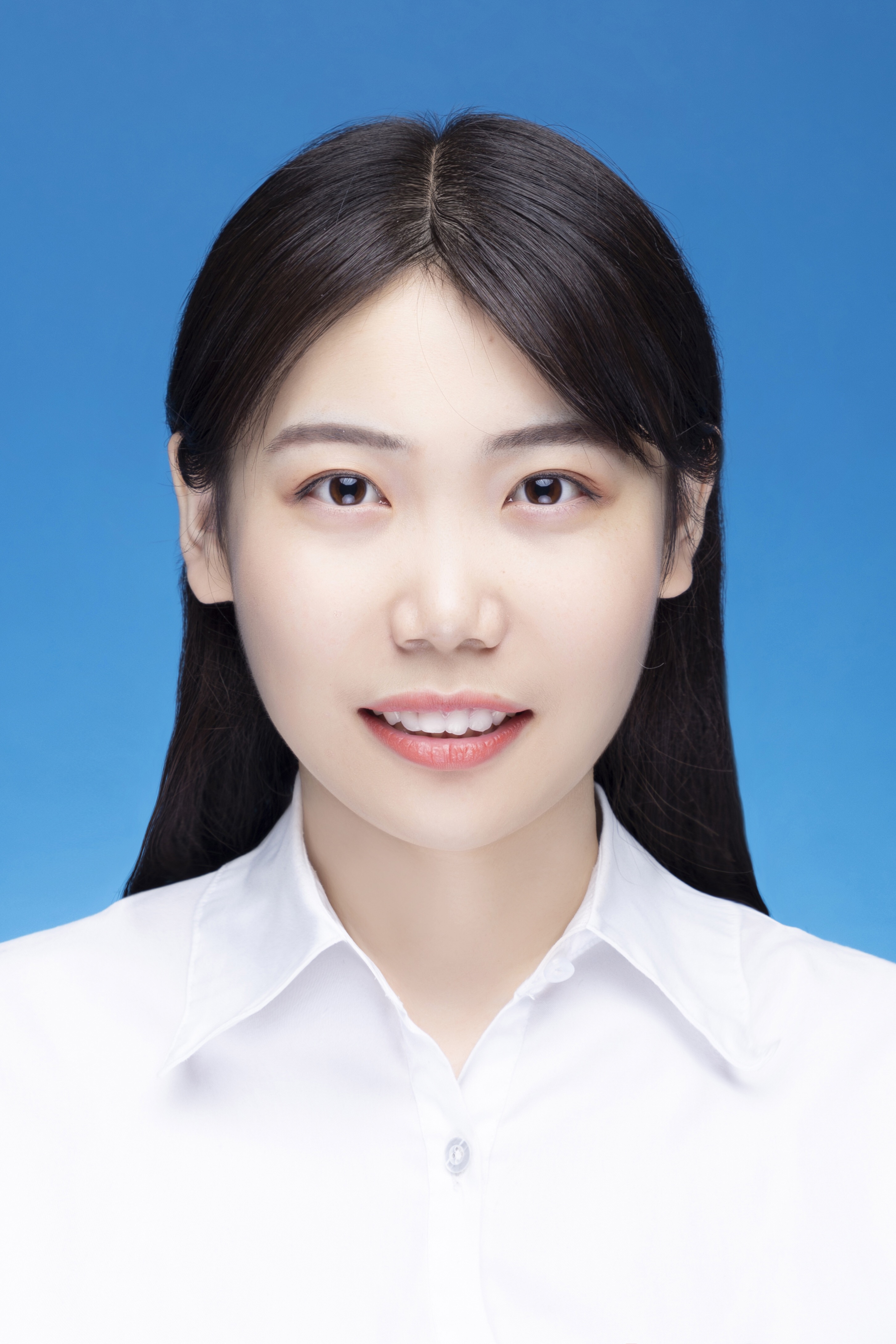}}]{Xin Su}
received the M.S. degree from Beijing Jiaotong University, China, in 2023. She is currently pursuing the Ph.D. degree at the School of Computer Science and Technology, Beijing Jiaotong University. Her research interests primarily focus on Multimodal Large Language models (MLLMs) and Multimodal Information Extraction.
\end{IEEEbiography}

\begin{IEEEbiography}[{\includegraphics[width=1in,height=1.25in,clip,keepaspectratio]{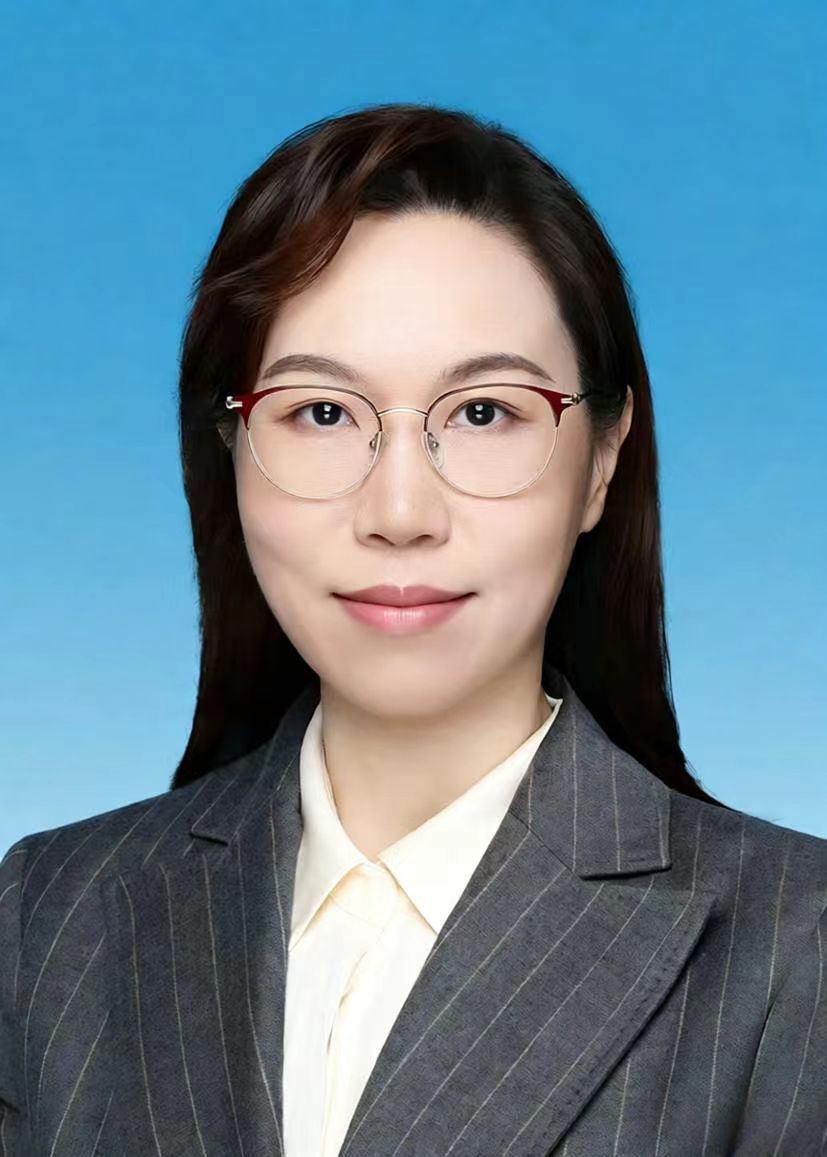}}]{Yi Jin}
(Member, IEEE) received the Ph.D. in signal and information processing from the Institute of Information Science, Beijing Jiaotong University, Beijing, China, in 2010. She was a Visiting Scholar with the School of Electrical and Electronic Engineering, Nanyang Technological University, Singapore, from 2013 to 2014. She is currently a Full Professor with the School of Computer Science and Technology, Beijing Jiaotong University. Her research interests include computer vision, pattern recognition, image processing, and machine learning.
\end{IEEEbiography}
\end{document}